%% file: main.tex
\pdfoutput=1
\documentclass[11pt]{article}

\usepackage[font=libertinus, citestyle=numeric]{kurbanlab}
\input{affiliations}

\usepackage[flushleft]{threeparttable}
\titleformat{\paragraph}[runin]
  {\normalfont\kilsemi\fontsize{9.8}{11}\selectfont\color{kilink}}{}{0pt}{}
\apptocmd{\appendix}{}{}{}

\newcommand{\pass}{\mathrm{Pass@1}}
\renewcommand{\Cov}{\mathrm{C}}
\newcommand{\rauc}{\mathrm{rAUC}}
\definecolor{cA}{RGB}{31,78,121}
\definecolor{cB}{RGB}{197,90,17}
\definecolor{cC}{RGB}{84,130,53}
\definecolor{cD}{RGB}{112,48,160}
\definecolor{cGrey}{RGB}{115,115,115}

\title{Oracle Gaps in Reliability Coverage:\\Sampling Noise or Policy Specialization?}
\Subtitle{Separating a winner's curse from real specialization before training a router}
\RunningTitle{Oracle gaps in reliability coverage}

\Author{Mert Onur Cakiroglu}{iub}
\Author{Mehmet Dalkilic}{iub}
\Author[corresponding=hkurban@hbku.edu.qa]{Hasan Kurban}{hbku}

\CodeURL{https://github.com/KurbanIntelligenceLab/oracle-gaps}

\begin{document}
\maketitle

\begin{abstract}
Policies trained from the same base model can appear to solve different problems. An oracle that chooses the best policy for each problem may therefore appear much stronger than any single policy. Selecting the largest estimated success rate also selects favorable sampling errors. We study this effect through reliability coverage, the fraction of problems whose success probability reaches a chosen threshold. Our first test redistributes stored correctness outcomes across policies within each problem. A second also preserves each policy's total successes, accounting for overall quality differences under a specified statistical model. For five training seeds of a seven-billion-parameter vision--language model, redistribution reproduces $0.096$ of an estimated $0.113$ oracle gap at threshold $0.10$. Neither test finds significant evidence at this threshold. Small advantages remain unresolved. Mixtures, routers, voting, and weight averaging show no detectable improvement over their corresponding single-policy baselines. Training policies on different datasets shows little detectable specialization under light post-training, and no router gain. A stronger recipe does create it, both tests detect it, and a router gain appears only at the high thresholds where the specialists separate. In a control with predictable specialization, a router recovers about half the oracle gap. Coverage bounds explain why even a genuine oracle advantage need not yield a deployment gain. The tests assess apparent specialization from stored responses before investment in~routing.\\[3pt]
Code: \url{https://github.com/KurbanIntelligenceLab/oracle-gaps}
\end{abstract}

{\raggedright\printkeywords}

\section{Introduction}

Training the same base model several times can produce policies that appear to solve different subsets of an evaluation set, even when only the random seed changes. Here, a \emph{policy} is the response-generating behavior of a trained model, and a \emph{pool} is a collection of such policies. We ask whether the apparent differences within a pool reflect stable strengths on different problems or noise from sampling a limited number of~\mbox{responses}.

A single correct response does not establish that a policy solves a problem reliably. We therefore choose a \emph{reliability threshold} $\tau$. A problem counts as covered when the policy's probability of answering correctly is at least $\tau$. The policy's \emph{coverage} is the fraction of problems that satisfy this requirement.\footnote{Some repeated-sampling work uses \emph{coverage} for the fraction of problems solved at least once in $k$ responses, or $\mathrm{Pass@}k$ \citep{brown2024monkeys}. Here, coverage refers to the reliability-threshold definition.} An idealized \emph{oracle} chooses the policy with the highest true success probability on each problem. Its advantage over the best single policy is the opportunity available to a perfect selector. In practice, the true probabilities are unknown. Substituting estimates from sampled responses gives a \emph{plug-in oracle}, whose apparent advantage may include sampling~\mbox{error}.

\begin{figure}[t]
\centering
\includegraphics[width=\textwidth]{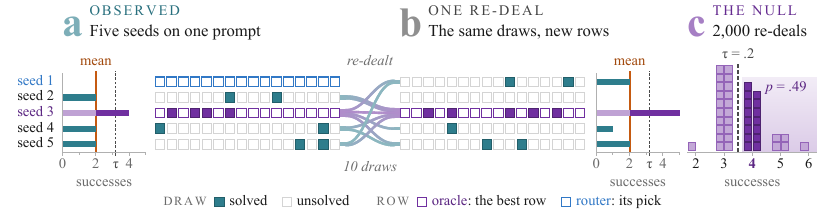}
\caption{\textbf{Constructing the re-deal null for one problem.} \textbf{(a)} Each row contains $16$ binary correctness outcomes from one of five policies. Averaging all $80$ outcomes estimates the success probability of choosing a policy uniformly for each response. \textbf{(b)} A re-deal randomly redistributes these outcomes into five groups of $16$, preserving all $10$ successes but removing the original policy assignments. \textbf{(c)} Distribution of the largest group success count over $2{,}000$ re-deals. The permutation $p$-value is the fraction with a maximum at least as large as the observed four successes. The dashed line marks the reliability threshold $\tau=0.2$, equivalent to $3.2$ successes out of $16$. An integer count must therefore reach four to clear~it.}
\label{fig:redeal}
\end{figure}

A deployed system must choose without knowing which policy will succeed. It can use one policy for every problem, \emph{mix} policies by sampling one for each response, or \emph{route} each problem using its text or image. On Geometry3K, a visual geometry benchmark, our five-policy plug-in oracle exceeds the best single policy's coverage by $0.113$ at $\tau=0.10$. Yet neither mixing nor the tested routers yields a detectable gain. Similar discrepancies have been reported for heterogeneous model pools~\citep{chen2026cofailure,li2026routerbench,kim2026diversityaudit}.

We examine two obstacles to recovering an oracle advantage. The first is a statistical effect known as the \emph{winner's curse}. Selecting the largest estimated rate favors positive estimation errors \citep{fowler2026capfrontier,chen2026routinggap}. Our first reference model, the \emph{re-deal null}, measures how large a gap this selection can produce when policies have equal success probabilities on each problem. It randomly redistributes stored correctness outcomes across policies within each problem, preserving the total successes (Figure~\ref{fig:redeal}). A second, \emph{margin-preserving null} also fixes each policy's total successes across problems, allowing systematic quality differences under a specified model. Simulations and deliberately introduced specialization assess both~\mbox{tests}.

The second obstacle arises when a deployment rule cannot recover a genuine oracle advantage. A mixture averages success probabilities and can fall below a threshold that one member exceeds. A router must identify a suitable member from the problem's available features. Here, \emph{specialization} means that members have stable relative strengths on different problems. Detecting such differences and predicting which member to use are separate tasks. Section~\ref{sec:setup} formalizes this distinction with a routing identity and a coverage guarantee for uniform~\mbox{mixing}.

Our primary experiment trains five policies from one seven-billion-parameter (7B) vision--language model, changing only the random seed. The runs are \emph{exchangeable by design}, meaning that permuting their seed labels leaves their joint distribution unchanged. No seed is favored in advance, although realized performance can differ \citep{fort2019landscape,henderson2018matters,dodge2020finetuning,wu2026sparsity,zhu2025pathnottaken}. We also test policies trained on different datasets, under the same light recipe and under a stronger one. Under light training the nulls reproduce much of the estimated gap in both settings and no tested deployment rule improves detectably. Stronger training creates specialization that both tests detect and that a router collects, if at all, only at high reliability thresholds. Positive controls show when the tests detect specialization and when a router can exploit~it.

\paragraph{Contributions.} We make three~\mbox{contributions}.
\begin{enumerate}[leftmargin=1.4em,itemsep=1pt,topsep=2pt]
\item \textbf{Tests of apparent specialization.} Two null models use stored responses to assess oracle gaps, with simulations and positive controls that characterize their false-positive rates and power (Section~\ref{sec:protocol},~Appendix~\ref{app:audit}).
\item \textbf{Bounds that turn a gap into a requirement.} A routing identity states the accuracy a router must reach on the prompts where members disagree to beat the best member, and a factor-$M$ bound gives the threshold shift a uniform mixture must survive. Every deployment comparison is read through them~(Section~\ref{sec:setup}).
\item \textbf{Controlled comparisons of policy pools.} We vary training seeds and training data, then evaluate mixtures, five router families, voting, and weight averaging against the corresponding single-policy baselines (Sections~\ref{sec:protocol}~and~\ref{sec:specialist}).

\end{enumerate}

\section{Reliability coverage and the limits of selection}
\label{sec:setup}

We first define reliability coverage at the population level, where each policy's success probabilities are known. We then describe how sampling affects its estimation. Let $\mathcal{P}$ be the space of evaluation prompts, including any associated image, and let $\mathcal{X}$ be their distribution. A policy $\pi$ assigns each prompt $x\in\mathcal{P}$ a response distribution $\pi(\cdot\mid x)$. For a sampled response $y$, let $v(x,y)$ equal $1$ when the answer is correct and $0$ otherwise. The policy's success probability~is
\[
p_\pi(x)=\Pr_{y\sim\pi(\cdot\mid x)}\!\left[v(x,y)=1\right].
\]
Following \citet{dragoi2025beyondpassk}, its coverage at reliability threshold $\tau$~is
\begin{equation}
\Cov_\pi(\tau)=\Pr_{x\sim\mathcal{X}}[\,p_\pi(x)\ge\tau\,],\qquad 0<\tau\le1.
\end{equation}
Increasing $\tau$ can only decrease coverage. For later bounds, we extend the definition by setting $\Cov_\pi(\tau)=0$ for $\tau>1$. In evaluation, we estimate $p_\pi(x)$ from $k$ sampled responses. The raw estimate is the number of correct responses divided by $k$. When $0<\tau\le1/k$, this estimate clears the threshold exactly when at least one response is correct. At these thresholds, empirical coverage therefore uses the same success event as~$\mathrm{Pass@}k$.

\paragraph{Summarizing a coverage curve.} Choose a lower threshold $\tau_{\mathrm{lo}}$ and an upper threshold $\tau_{\mathrm{hi}}$, with $0<\tau_{\mathrm{lo}}<\tau_{\mathrm{hi}}\le1$. Over this \emph{reliability band}, define the \emph{restricted area under the coverage curve} (rAUC)~as
\begin{equation}\label{eq:rauc}
\rauc(\pi)=
\frac{1}{\tau_{\mathrm{hi}}-\tau_{\mathrm{lo}}}
\int_{\tau_{\mathrm{lo}}}^{\tau_{\mathrm{hi}}}
\Cov_\pi(\tau)\,d\tau.
\end{equation}
The normalization makes rAUC the average coverage over the band. Integrating over the full range $(0,1]$ gives $\mathbb{E}_{x\sim\mathcal{X}}[p_\pi(x)]$, the probability that one response to a random prompt is correct. We call this quantity $\mathrm{Pass@1}$, or average accuracy~\citep{dragoi2025beyondpassk}.

For Geometry3K, we fixed the band $[0.05,0.30]$ and reporting thresholds $\{0.05,0.10,0.20,0.30\}$ before scoring the test split. The primary evaluation uses $k=16$ responses per policy and prompt. With raw frequencies, $\tau=0.05$ requires at least one correct response, while the headline threshold $\tau=0.10$ requires at least two. Appendix~\ref{app:protocol} examines sensitivity to the band and the success-rate~\mbox{estimator}.

\paragraph{Policies and deployment rules.} Let $B$ denote the base model's policy before post-training, and let $P_1,\dots,P_M$ be a pool of $M$ policies trained from that model. Section~\ref{sec:protocol} varies only the training seed. Section~\ref{sec:specialist} also varies the training data. A router $r:\mathcal{P}\to[M]$, where $[M]=\{1,\dots,M\}$, selects a policy from the prompt. With probabilities taken over $x\sim\mathcal{X}$ and the argument $x$ omitted where unambiguous,~\mbox{define}
\begin{align}
\Cov_{\mathrm{orac}}(\tau)&=\Pr[\textstyle\max_m p_{P_m}\ge\tau], &
\Cov_{\mathrm{best}}(\tau)&=\textstyle\max_m\Pr[p_{P_m}\ge\tau],\\
\Cov_{\mathrm{mix}}(\tau)&=\Pr[\textstyle\frac1M\sum_m p_{P_m}\ge\tau], &
\Cov_{r}(\tau)&=\Pr[p_{P_{r(x)}}(x)\ge\tau]
\end{align}
for the oracle, best single member, uniform mixture, and router, respectively. The best single member maximizes coverage at the chosen threshold and is then used on every prompt. The uniform mixture draws a member with probability $1/M$ for each~\mbox{response}.

The oracle's advantage over the best single member is the \emph{population oracle~gap},
\[
L(\tau)=\Cov_{\mathrm{orac}}(\tau)-\Cov_{\mathrm{best}}(\tau).
\]
This gap is latent because it depends on unknown success probabilities. Experiments use \emph{plug-in estimates}, replacing those probabilities with estimated rates. Numerical results use the population notation, and we say \emph{population} or use hats only where both appear together, for the gap, disagreement, and routing error alike. The theoretical best member maximizes coverage at each threshold. Comparisons that select one member by rAUC are identified~\mbox{separately}.

\subsection{Where member choice matters}
\label{sec:calculus}

A selector can change coverage only on prompts where some members meet the threshold and others do not. Mixing also leaves coverage unchanged when all members lie on the same side of the threshold, because their average stays on that~\mbox{side}.

\begin{definition}[Disagreement set]
The \emph{disagreement set}~is
\[
\mathcal{D}(\tau)=
\left\{x:\max_m p_{P_m}(x)\ge\tau>\min_m p_{P_m}(x)\right\}.
\]
It contains the prompts on which at least one member meets $\tau$ and at least one falls below it. Its~\mbox{probability},
$D(\tau)=\Pr_{x\sim\mathcal{X}}[x\in\mathcal{D}(\tau)]$, is the fraction of prompts on which member choice can affect~\mbox{coverage}.
\end{definition}

The oracle's coverage is a ceiling for selection and mixing because neither can exceed the largest member rate on a prompt. It can outperform the best single member only on the disagreement set, so $L(\tau)\le D(\tau)$. The key question is how much of that opportunity a router loses through incorrect~\mbox{selection}.

\begin{proposition}[Routing loss and break-even accuracy]
\label{thm:routing}
For any router~$r$,
\begin{equation}
\begin{aligned}
\Cov_{\mathrm{orac}}(\tau)-\Cov_{r}(\tau)
&=\Pr\!\left[\max_m p_{P_m}\ge\tau,\ p_{P_{r(x)}}(x)<\tau\right]\\
&=D(\tau)e_D(r,\tau),
\end{aligned}
\end{equation}
where
\[
e_D(r,\tau)
=\Pr\!\left[p_{P_{r(x)}}(x)<\tau\mid x\in\mathcal{D}(\tau)\right]
\]
is the probability of selecting a member below threshold, conditional on a disagreement prompt. Set $e_D=0$ when $D(\tau)=0$. For $D(\tau)>0$, the router beats the best single member exactly when $e_D(r,\tau)<L(\tau)/D(\tau)$. When $D(\tau)=0$, every member has the same coverage decisions and routing cannot change~\mbox{coverage}.
\end{proposition}

Equivalently, the router's gain is $L(\tau)-D(\tau)e_D(r,\tau)$. Reporting both $D$ and $e_D$ distinguishes a small opportunity from inaccurate selection where member choice~\mbox{matters}.

\subsection{What a mixture can guarantee}
\label{sec:factorM}

A uniform mixture averages all members' success probabilities. The following result states what this averaging can~\mbox{guarantee}.

\begin{proposition}[Factor-$M$ threshold shift]
\label{thm:factorM}
For every pool and every~$\tau>0$,
\begin{equation}
\Cov_{\mathrm{mix}}(\tau)\;\ge\;\Cov_{\mathrm{orac}}(M\tau).
\end{equation}
The right-hand side is zero when $M\tau>1$. The factor $M$ cannot be reduced in general. Equality holds whenever at most one member has nonzero success probability on each prompt. Within this family, every replacement factor $c$ with $0<c<M$ fails for some pool and~\mbox{threshold}.
\end{proposition}

To see the bound, note that a member with success probability at least $M\tau$ raises the average of $M$ nonnegative rates to at least $\tau$. For example, with two members and $\tau=0.10$, the guarantee compares mixture coverage at $0.10$ with oracle coverage at $0.20$. If several members succeed on a prompt, the mixture can lie well above the~\mbox{bound}.

\section{Do training seeds produce a recoverable oracle advantage?}
\label{sec:protocol}

We begin with a pool whose members share the same base model and training procedure. This experiment asks whether changing the training seed produces stable differences that a deployment rule can exploit. All analyses use stored sampled responses, which we call~\emph{rollouts}.

\subsection{Experimental design}

\paragraph{Models and data.} We train five policies from Qwen2.5-VL-7B-Instruct \citep{bai2025qwen25vl} on the same $400$ Geometry3K prompts \citep{lu2021intergps}, changing only the random seed. Each run uses $500$ steps of Group Relative Policy Optimization (GRPO), a reinforcement-learning method that compares rewards among responses to the same prompt. Rewards come from an automatic correctness checker \citep{shao2024grpo,guo2025deepseekr1}. The primary test set contains $240$ prompts, with $16$ responses per policy and prompt. We also repeat the study on MathVista, a visual mathematical reasoning dataset \citep{lu2024mathvista}. Its selection and reliability band follow a rule fixed before test scoring (Appendix~\ref{app:ds2}). All reported results use the corrected verifier described in~Appendix~\ref{app:integrity}.

\paragraph{What the design controls.} The five Geometry3K policies have test mean success rates between $0.355$ and $0.376$. On validation, one seed leads by about $0.07$ coverage at low thresholds. Such a leader is compatible with exchangeability of the training procedure and motivates testing both a null with equal member rates and one that preserves overall member performance. Appendix~\ref{app:protocol} reports training details, pairwise similarity, and uncertainty~\mbox{estimates}.

\subsection{How much of the estimated gap does sampling explain?}

The estimated rates suggest substantial room for selection. At $\tau=0.10$, members disagree about coverage on $D=0.233$ of test prompts. The plug-in oracle gap is $L=0.113$ (Table~\ref{tab:null}). Their ratio, $L/D=0.482$ from unrounded values, means that a router must choose a threshold-clearing member on more than $51.8\%$ of disagreement prompts to beat the best fixed member (Proposition~\ref{thm:routing}). We first test how much of this estimated gap sampling alone can~\mbox{produce}.

\begin{table}[!ht]
\centering
\caption{\textbf{Oracle-gap tests on $240$ Geometry3K test prompts.} Each policy has $16$ responses per prompt. Estimates use raw frequencies. \emph{Obs.} is the observed statistic and \emph{null} the plain-null mean. \emph{Null $90\%$} is the $5$th--$95$th percentile range of randomized statistics, not an interval for the population gap. The upper-tail fractions $p$ and $p_{\mathrm{m}}$ use the plain and margin-preserving nulls, respectively. \emph{Split-half} selects both per-prompt members and the fixed comparator on eight responses and scores them on the other eight. Re-dealing preserves the uniform mixture's raw~rate.}
\label{tab:null}
\scriptsize
\setlength{\tabcolsep}{2.0pt}
\begin{tabular}{@{}l S[table-format=1.3] S[table-format=1.3] S[table-format=1.3] S[table-format=1.3] c S[table-format=1.3] S[table-format=1.3] S[table-format=+1.3] S[table-format=1.3] S[table-format=1.3] S[table-format=1.2] S[table-format=+1.3] S[table-format=+1.3] S[table-format=1.2]@{}}
\toprule
& \multicolumn{2}{c}{\textbf{$D(\tau)$}} & \multicolumn{6}{c}{\textbf{$L(\tau)$}} & \multicolumn{3}{c}{\textbf{Oracle $-$ mixture}} & \multicolumn{3}{c}{\textbf{Rise in $L$, $M{=}2$ to $5$}} \\
\cmidrule(lr){2-3}\cmidrule(lr){4-9}\cmidrule(lr){10-12}\cmidrule(lr){13-15}
\textbf{$\tau$} & {\bfseries obs.} & {\bfseries null} & {\bfseries obs.} & {\bfseries null} & {\bfseries null $90\%$} & {\bfseries $p$} & {\bfseries $p_{\mathrm{m}}$} & {\bfseries split-half} & {\bfseries obs.} & {\bfseries null} & {\bfseries $p$} & {\bfseries obs.} & {\bfseries null} & {\bfseries $p$} \\
\midrule
$0.05$ & 0.296 & 0.277 & 0.142 & 0.125 & $[0.108,0.137]$ & 0.036 & 0.027 & +0.021 & 0.150 & 0.150 & 1.00 & +0.076 & +0.069 & 0.11 \\
$0.10$ & 0.233 & 0.220 & 0.113 & 0.096 & $[0.079,0.113]$ & 0.074 & 0.063 & +0.020 & 0.113 & 0.104 & 0.22 & +0.058 & +0.051 & 0.19 \\
$0.20$ & 0.208 & 0.196 & 0.075 & 0.078 & $[0.062,0.096]$ & 0.67  & 0.651 & +0.001 & 0.071 & 0.059 & 0.14 & +0.034 & +0.038 & 0.69 \\
$0.30$ & 0.196 & 0.193 & 0.075 & 0.080 & $[0.062,0.096]$ & 0.75  & 0.714 & -0.002 & 0.104 & 0.104 & 0.60 & +0.037 & +0.040 & 0.67 \\
\bottomrule
\end{tabular}
\end{table}

\paragraph{Two reference models.} Both nulls assume independent response draws. Under the plain null, all members have the same true success probability on each prompt. We randomly redistribute that prompt's $80$ binary outcomes into five groups of $16$. This preserves the total successes and the uniform mixture's raw~\mbox{rate}.

The margin-preserving null also fixes each member's total successes across prompts. It allows systematic quality differences through an additive model. If $q_{im}$ is member $m$'s success probability on prompt $i$, the model~is
\[
\operatorname{logit}(q_{im})=a_i+b_m,
\qquad \operatorname{logit}(q)=\log\!\frac{q}{1-q}.
\]
Here $a_i$ describes prompt difficulty and $b_m$ describes member quality on the log-odds scale. A separate prompt-by-member effect would violate the~\mbox{model}.

Let $c_{im}$ count successes in $k$ responses, with $n$ prompts and $M$ members. Conditioning on both prompt and member totals removes the unknown parameters $(a,b)$. Each feasible count table then has probability proportional~to
\[
\prod_{i=1}^{n}\prod_{m=1}^{M}\binom{k}{c_{im}}.
\]
The binomial weights count the response assignments represented by each table. We sample $2{,}000$ count tables using random transfers that preserve both totals, with acceptance probabilities chosen to target these weights. Appendix~\ref{app:audit} gives the derivation, sampling procedure, and convergence~\mbox{checks}.

\paragraph{Observed gaps and null expectations.} At $\tau=0.10$, the plain-null mean is $0.096$, compared with observed gap $0.113$. We call their difference, $+0.017$, the \emph{excess}. A $95\%$ bootstrap percentile interval is $[-0.016,+0.025]$, obtained by repeatedly resampling prompts and recomputing the null. Neither test rejects at level $0.05$ ($p=0.074$, $p_{\mathrm{m}}=0.063$). At $\tau=0.05$, the excess is about $+0.017$ and the plain test rejects before adjustment for multiple tests ($p=0.036$). At $\tau\ge0.20$, the gap lies below the null~\mbox{mean}.

\paragraph{Independent selection and scoring.} We also split each member's responses in half, select a member for each prompt using eight responses, and score it on the remaining eight. The fixed comparator is selected on the first half too. The resulting gaps are $+0.021$ and $+0.020$ at $\tau=0.05$ and $0.10$, and within $0.002$ of zero above (Table~\ref{tab:null}). This separates selection from scoring, though the estimates still carry sampling noise on a coarser eight-response~\mbox{grid}.

\paragraph{What the tests can detect.} Null simulations yield rejection rates of $3\%$--$6\%$ at nominal level $5\%$. To assess sensitivity, we plant specialization by transferring successes within each prompt to a designated member. With a target of one transfer per prompt, the tests reject in $65\%$ and $70\%$ of $20$ plantings. With two, both reject in every~\mbox{planting}.

We also simulate diffuse differences by independently perturbing every member's log-odds on each prompt (Table~\ref{tab:diffuse}). At a population gap near $0.05$, each test detects only half the simulated alternatives. Increasing the response budget from $16$ to $64$ raises that power to $90\%$ (Appendix~\ref{app:audit}). Non-rejection at the primary budget therefore leaves meaningful advantages~\mbox{unresolved}.

\begin{table}[!htbp]
\centering
\caption{\textbf{Power against diffuse rate differences at $k=16$.} Independent Gaussian perturbations with standard deviation $\sigma$ are added to every member's log-odds of success, centered on the observed prompt mean. No sole specialist is designated. The table reports the population gap, excess over the plain-null mean, and rejection fractions at level $0.05$ over $20$~replicates.}
\label{tab:diffuse}
\small
\setlength{\tabcolsep}{6pt}
\begin{tabular}{@{}S[table-format=1.2] S[table-format=1.3] S[table-format=+1.3] S[table-format=1.2] S[table-format=1.2]@{}}
\toprule
{$\sigma$} & {$L$} & {excess} & {power, re-deal} & {power, margin} \\
\midrule
0.25 & 0.023 & +0.003 & 0.00 & 0.05 \\
0.50 & 0.052 & +0.013 & 0.50 & 0.50 \\
1.00 & 0.108 & +0.052 & 1.00 & 1.00 \\
\bottomrule
\end{tabular}
\end{table}

\paragraph{When detected differences support routing.} Combining five 7B policies with five from a three-billion-parameter (3B) model makes the margin-preserving test reject at $\tau\le0.10$ ($p<0.001$). The tested routers recover no gain. In another control, the specialist is assigned by a fixed prompt feature. With a target of two transfers, both tests reject. A router using similar validation prompts recovers $+0.079$ of the $0.146$ oracle gap at $\tau=0.10$. Appendix~\ref{app:audit} describes the controls and checks sensitivity to the null~\mbox{model}.

\paragraph{More responses and the population gap.} With $64$ fresh responses per member, the Geometry3K gap at $\tau=0.10$ falls to $0.058$, with null mean $0.047$ and excess $+0.012$ from unrounded values ($p=0.12$). The excess can substantially understate the population gap. In simulations at $k=16$, it recovers only $14\%$--$35\%$ of population gaps of at least $0.02$~(Appendix~\ref{app:audit}).

\paragraph{Multiple tests.} Of $156$ tests across the light-recipe seed and cross-dataset pools and their replications, four survive Benjamini--Hochberg correction, which targets a false-discovery rate of $0.05$ (adjusted $p=0.039$ each). All concern cross-dataset pools: two oracle-gap tests and one disagreement test for ten members at $\tau=0.05$, and two-member disagreement at $\tau=0.10$. No seed-pool result survives, and the stronger-recipe pool of Section~\ref{sec:specialist} rejects with no null draw in $2{,}000$ at or above its observed~gap.

\subsection{Can deployment rules recover a gain?}

We compare deployment rules at the same response budget. For the uniform mixture, we subsample $16$ responses from the pooled $80$ on each prompt and average over $200$ subsamples. This gives the mixture the same response count and threshold grid as a single~\mbox{member}.

\paragraph{Mixing and weight averaging.} The plug-in oracle gains $0.095$ rAUC over the best member on Geometry3K and $0.069$ on MathVista (Table~\ref{tab:main}). The Geometry3K mixture reaches $0.592$ rAUC, compared with $0.607$ for the member selected on test. The difference is $-0.015$ with $95\%$ interval $[-0.033,+0.001]$. Against the member selected on validation, the difference is $0.000$ with interval $[-0.020,+0.015]$. Validation selection is the deployable comparison, since test outcomes are unavailable when choosing a~\mbox{policy}.

Averaging the five policies' parameter updates produces one policy, often called a \emph{model soup}. The averaged policy shows no detectable gain over the best member, and its rAUC is within $0.004$ of the budget-matched mixture on every split~(Appendix~\ref{app:protocol}).

\begin{table}[!htbp]
\centering
\caption{\textbf{Policy selection and mixing on the two test sets.} Both evaluations use $k=16$ responses per policy and prompt, and rates are posterior means (Appendix~\ref{app:protocol}). Geometry3K has $240$ test prompts and reliability band $[0.05,0.30]$. MathVista has $184$ prompts and band $[0.50,0.90]$. The best member maximizes test rAUC. The plug-in oracle chooses a member separately for each prompt using estimated success rates. The mixture is scored by subsampling $16$ responses from the pool. The soup is one policy formed by averaging the five parameter updates. The routing row uses the family selected on~validation.}
\label{tab:main}
\small
\setlength{\tabcolsep}{3pt}
\begin{tabular}{@{}l l S[table-format=1.3] S[table-format=1.3] S[table-format=1.3] S[table-format=1.3]@{}}
\toprule
\textbf{Rule} & \textbf{System} & {\bfseries Geometry3K $\rauc$} & {\bfseries $\pass$} & {\bfseries MathVista $\rauc$} & {\bfseries $\pass$} \\
\midrule
single    & base $B$                                   & 0.579 & 0.354 & 0.498 & 0.610 \\
single    & best member by test rAUC                  & 0.607 & 0.376 & 0.543 & 0.632 \\
\midrule
oracle    & best member per prompt                     & 0.702 & 0.451 & 0.612 & 0.698 \\
mixture   & uniform mixture of the seeds              & 0.592 & 0.368 & 0.534 & 0.628 \\
soup      & mean of the five updates                   & 0.592 & 0.369 & 0.536 & 0.622 \\
routing   & best family on validation                  & 0.603 & 0.374 & 0.533 & 0.631 \\
\bottomrule
\end{tabular}
\end{table}

\paragraph{Routing from prompt features.} We train three text-based router families on validation outcomes: a linear classifier, gradient-boosted decision trees, and a nearest-neighbor rule. At $\tau=0.10$, their disagreement errors are $0.518$, $0.500$, and $0.411$, compared with the estimated break-even error $0.482$. Their coverage gains are $-0.008$, $-0.004$, and $+0.017$. Every interval includes zero (Table~\ref{tab:routing}). At $\tau\ge0.20$, all three trail the best member. Two additional families use vector representations, or \emph{embeddings}, of both the image and text from the base model. Their gains across both datasets range from $-0.037$ to $0.000$, again with no interval wholly above zero. Appendix~\ref{app:protocol} specifies all five~\mbox{families}.

\subsection{Voting and robustness checks}

\paragraph{Voting.} A vote combines multiple responses into one answer and can improve success probability even when selecting a member cannot. We therefore compare pooled voting with \emph{self-consistency}, which votes over the same number of responses from one policy. At nine responses per prompt, the Geometry3K pooled vote gains $0.011$ rAUC over the best member's own vote, with interval $[-0.013,+0.035]$. On MathVista, it trails that vote by $0.015$~(Appendix~\ref{app:vote}).

\paragraph{Pool size.} Increasing the number of members gives the plug-in oracle more noisy estimates to choose from. We average each statistic over all subsets of a given size and apply the same comparison after re-dealing. On Geometry3K at $\tau=0.10$, increasing the pool from two to five members raises the oracle-minus-best-member gap by $0.058$, compared with a null increase of $0.051$ ($p=0.19$; Table~\ref{tab:null}). The larger apparent opportunity is therefore consistent with the increase expected from selection~\mbox{alone}.

\paragraph{More responses, more prompts, and a smaller model.} Three prespecified extensions increase the evaluation budget to $64$ responses, add $361$ Geometry3K test prompts, and repeat training with five 3B seeds (Appendix~\ref{app:replications}). None of the $25$ coverage-gap tests rejects the plain re-deal null at level $0.05$. Matched split-half gaps remain within $\pm0.015$, and none of $63$ routing intervals lies wholly above zero. The budget-matched mixture trails the best member throughout, at $k=64$ with both intervals below~\mbox{zero}.

\section{Does training on different datasets create specialists?}
\label{sec:specialist}

\begin{figure}[!htbp]
\centering
\includegraphics[width=\textwidth]{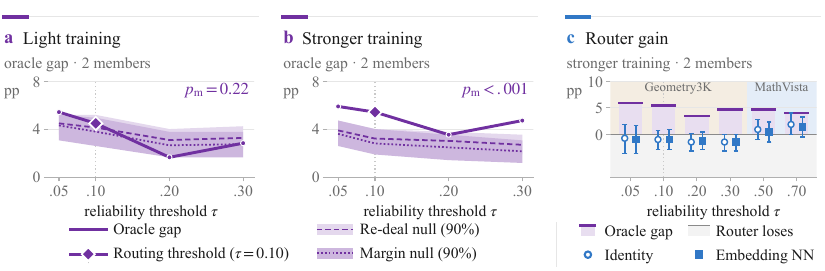}
\caption{\textbf{Cross-dataset pools under light and stronger training.} \textbf{(a,b)} Two-member pools, one seed trained on each dataset, on the $424$ union test prompts with $16$ responses per member and prompt. Solid curves are the plug-in oracle gap, dashed and dotted curves the plain and margin-preserving null means, shaded over the central $90\%$ of each null distribution. Diamonds mark $\tau=0.10$, where $p_{\mathrm m}$ reports the margin-preserving test. \textbf{(c)} The stronger two-member pool at every reported threshold: oracle gap (purple bars) and router gains over the best member (blue points, $95\%$ prompt-bootstrap intervals) for the dataset-identity router and the embedding nearest-neighbor (NN) router, with each dataset's reliability band shaded and the grey region marking a router below the best member. All vertical values are percentage~points.}
\label{fig:training}
\end{figure}

Different training data might create complementary strengths even when changing the seed does not. We test this under the light recipe of Section~\ref{sec:protocol} and under a stronger one, $1{,}500$ GRPO steps on $1{,}200$ Geometry3K or all $366$ MathVista training prompts with rank-$64$ adapters (Appendix~\ref{app:protocol}). Under each recipe we combine policies trained on Geometry3K with policies trained on MathVista and score every member on the union of the two test sets, $424$ prompts. The primary pools hold one seed from each training set (Figure~\ref{fig:training}). Appendix~\ref{app:specialist} reports all ten light seeds and all four stronger~\mbox{seeds}.

Light training made no specialists that the tests resolve above $\tau=0.05$ or that a router could use. The Geometry3K-trained member has higher average accuracy on both datasets, $0.377$ and $0.628$ against $0.346$ and $0.621$, and at $\tau=0.10$ it also covers more prompts in both halves. The oracle gap at $\tau=0.10$ is $0.045$ against plain and margin-preserving means of $0.041$ and $0.038$ ($p=0.35$ and $0.22$; Figure~\ref{fig:training}a and Table~\ref{tab:specialist}), and the excess over either null is at most $0.011$ at any threshold. A \emph{dataset-identity router}, which sends each prompt to the member trained on its dataset, gains $-0.005$ with disagreement error $0.41$ against break-even $0.37$, and both embedding routers gain $-0.005$ (Table~\ref{tab:specialist}). All ten light seeds read the same way (Appendix~\ref{app:specialist}). Disagreement itself exceeds the plain null ($D=0.120$ against $0.095$, $p<0.001$), so excess disagreement and a recoverable coverage advantage are distinct~\mbox{findings}.

Stronger training made specialists. Each member now leads on its own dataset: mean success rates are $0.452$ and $0.644$ for the Geometry3K-trained member and $0.388$ and $0.678$ for the MathVista-trained member, on Geometry3K and MathVista prompts respectively, so each gains $0.07$ to $0.10$ over the base model on its own dataset and about $0.035$ on the other. Both tests detect it. At $\tau=0.10$ the oracle gap is $0.054$ against null means of $0.032$ and $0.028$ ($p<0.001$ for both; Figure~\ref{fig:training}b and Table~\ref{tab:specialist_long}), an excess of $0.022$ and $0.026$, and the gap exceeds both nulls at every reported threshold except $0.20$ ($p=0.23$, $p_{\mathrm{m}}=0.063$). Disagreement rises to $D=0.130$ against a null of $0.075$. With all four stronger seeds the gap at $\tau=0.10$ is $0.092$ against $0.062$ and $0.057$~($p<0.001$).

Whether a router collects depends on the threshold. At $\tau=0.10$ none does: the identity router gains $-0.009$ with interval $[-0.028,+0.007]$ and disagreement error $0.49$ against break-even $0.42$, and the embedding routers gain $-0.009$ and $-0.002$ (Figure~\ref{fig:training}c and Table~\ref{tab:specialist_long}). Two successes in $16$ is a low bar, and the Geometry3K-trained member, the stronger overall, still clears it on more prompts of both datasets. The MathVista-trained member leads on MathVista prompts only from $\tau=0.50$, inside that dataset's band (Appendix~\ref{app:specialist}). At $\tau=0.70$ the identity router gains $+0.019$ with interval $[0.000,+0.040]$ and error $0.18$ against break-even $0.33$ (Figure~\ref{fig:training}c), and on the four-member pool the embedding nearest-neighbor router gains $+0.038$ with interval $[+0.005,+0.068]$ and error $0.22$ against $0.39$ (Table~\ref{tab:specialist_long}). That is one interval above zero among $36$ unadjusted comparisons, an exploratory reading. At $\tau=0.70$ all six router estimates on the two pools are positive, from $+0.007$ to $+0.038$. The tests flagged this pool from cached rollouts before any router was fitted, and the gain appears only where the specialists~\mbox{separate}.

\section{Related work}
\label{sec:related}

\paragraph{Diversity and model combination.} Ensemble theory relates predictive gains to member diversity \citep{kuncheva2004combining,kuncheva2003diversity,krogh1995ensembles,wood2023diversity,ali2026diversitylaw}. For language-model pools, \citet{chen2026cofailure} bounds selection accuracy by the fraction of queries on which some member is correct. Our bounds concern thresholded success probabilities. Independent seeds are a standard source of diversity \citep{lakshminarayanan2017ensembles,fort2019landscape}. \citet{li2025selfmoa} motivate comparing pooled responses with repeated sampling from the best model. Work on averaging fine-tuned weights motivates our model-soup comparison~\citep{wortsman2022soups,ilharco2023taskarithmetic}.

\paragraph{Reliability and oracle estimation.} Our coverage metric follows \citet{dragoi2025beyondpassk}. \citet{fowler2026capfrontier} study bias from maximizing noisy estimates. \citet{chen2026routinggap} uses seed-aligned responses to separate reproducible specialization from single-draw noise on heterogeneous text pools. Our tests require neither aligned response seeds nor additional samples. The margin-preserving test accounts for quality under an additive-logit model. It connects to conditional tests for differential item functioning, which assess performance differences across items beyond a shared difficulty model \citep{rasch1960probabilistic,holland1988dif,polo2024tinybenchmarks}. Routing benchmarks provide the setting for oracle comparisons \citep{hu2024routerbench,ong2024routellm}. \citet{miller2024errorbars} addresses uncertainty in single-model~\mbox{evaluation}.

\section{Discussion and limitations}

The tests are meant to run before any router is trained. Evaluate every member with the same response count on the same prompts. Report the estimated gap, disagreement, both null means, permutation fractions, and a bootstrap interval for the excess. Check power against alternatives that represent practically useful gains. A significant result motivates testing whether a router can predict a suitable member. Non-rejection is inconclusive when power is low. A router must achieve disagreement error below $L/D$ to beat the best member, using estimates at the same threshold and against the same~\mbox{comparator}.

The experiments post-train a shared base model, lightly in most pools and more strongly in one. Including replications, they cover two model sizes from one family, two datasets, and two training recipes. Substantially different model families and training regimes remain untested. Our routing results cover five lightweight families. Routers implemented by a language model remain~\mbox{untested}.

The margin-preserving test depends on the log-odds model. A member-quality effect that is constant on another scale can trigger rejection. In a simulation with constant effects on the probability scale, the test rejects in $33\%$ of replicates at $\tau=0.20$ (Appendix~\ref{app:audit}). Power also depends on the form of specialization and the response budget, as Table~\ref{tab:diffuse} shows. Bootstrap intervals measure prompt-sampling uncertainty for the trained policies, excluding variability from~\mbox{retraining}.

\section{Conclusion}

Large plug-in oracle gaps need not imply large gains from combining policies. Under light post-training, the re-deal nulls reproduce much of the estimated gap in our seed and cross-dataset pools, and no tested deployment rule yields a statistically resolved improvement over its single-policy baseline. Stronger training on different datasets creates specialization that both tests detect, and routers show a gain only at the thresholds where the specialists separate, as the positive controls do where member strengths are predictable from prompt features. Small or diffuse advantages remain possible. Tests based on stored responses, accompanied by power checks, can help decide when an apparent oracle advantage warrants further investment in~\mbox{routing}.

\begin{conflicts}
The authors declare no competing~\mbox{interests}.
\end{conflicts}

\bibliography{references}

\appendix

\setlength{\abovedisplayskip}{5pt plus 1pt minus 1pt}
\setlength{\belowdisplayskip}{5pt plus 1pt minus 1pt}
\setlength{\abovedisplayshortskip}{3pt plus 1pt minus 1pt}
\setlength{\belowdisplayshortskip}{3pt plus 1pt minus 1pt}

The appendices provide proofs (Appendix~\ref{app:proofs}), training and estimation details (Appendix~\ref{app:protocol}), and the statistical audit of oracle gaps (Appendix~\ref{app:audit}). They then describe the cross-dataset pools, voting, replications, and the verifier~\mbox{audit}.

\section{Proofs}
\label{app:proofs}

Throughout the proofs, $p_m(x)$ abbreviates $p_{P_m}(x)$. We omit $x$ when the meaning is clear. Unless stated otherwise, probabilities are over $x\sim\mathcal{X}$, and a fixed rule breaks ties between \mbox{maximizing}~\mbox{indices}.

\paragraph{Proposition~\ref{thm:routing}.} If $D(\tau)=0$, member choice cannot affect coverage, and both sides of the identity are zero. Now assume $D(\tau)>0$. Recall that $e_D$ is the conditional probability of choosing a member below threshold on a disagreement prompt. Choosing any member that meets $\tau$ preserves oracle coverage. Since the selected rate is pointwise at most the largest member~\mbox{rate},
\begin{align*}
\Cov_{\mathrm{orac}}(\tau)-\Cov_r(\tau)
&=\Pr\!\left[\max_m p_m\ge\tau\right]
-\Pr\!\left[p_{P_{r(x)}}\ge\tau\right]\\
&=\Pr\!\left[\max_m p_m\ge\tau,\ p_{P_{r(x)}}<\tau\right]\\
&=D(\tau)\Pr\!\left[p_{P_{r(x)}}<\tau\mid x\in\mathcal{D}(\tau)\right]\\
&=D(\tau)e_D(r,\tau).
\end{align*}
The loss event in the second line can occur only on the disagreement set, giving the conditional-probability factorization.~\mbox{Finally},
\begin{align*}
\Cov_r(\tau)>\Cov_{\mathrm{best}}(\tau)
&\quad\Longleftrightarrow\quad
\Cov_{\mathrm{orac}}(\tau)-D(\tau)e_D(r,\tau)
>\Cov_{\mathrm{orac}}(\tau)-L(\tau)\\
&\quad\Longleftrightarrow\quad
e_D(r,\tau)<\frac{L(\tau)}{D(\tau)}.
\end{align*}
Equivalently, the probability of choosing a threshold-clearing member on $\mathcal{D}(\tau)$ must exceed $1-L(\tau)/D(\tau)$. This is the break-even selection accuracy used in Section~\ref{sec:protocol}.~\qed

\paragraph{Proposition~\ref{thm:factorM}.} \emph{Lower bound.} If $\max_m p_m(x)\ge M\tau$, then nonnegativity~\mbox{gives}
\[
\frac1M\sum_{m=1}^M p_m(x)
\ge \frac1M\max_m p_m(x)
\ge \tau.
\]
Therefore
\[
\left\{\max_m p_m\ge M\tau\right\}
\subseteq
\left\{\frac1M\sum_{m=1}^M p_m\ge\tau\right\},
\]
and taking probabilities proves the bound. If $M\tau>1$, the left event is empty because~$p_m\le1$.

\emph{Tightness.} Suppose at most one member has nonzero success probability on each prompt. Then the sum of member rates equals their maximum,~\mbox{giving}
\[
\frac1M\sum_{m=1}^M p_m(x)\ge\tau
\quad\Longleftrightarrow\quad
\max_m p_m(x)\ge M\tau.
\]
The coverage events are identical, so the bound holds with~\mbox{equality}.

\emph{Why the factor cannot be reduced.} Let $h(x)=\max_m p_m(x)$ denote the only potentially nonzero rate.~\mbox{Then}
\[
\Cov_{\mathrm{mix}}(\tau)=\Pr[h\ge M\tau],
\qquad
\Cov_{\mathrm{orac}}(c\tau)=\Pr[h\ge c\tau].
\]
Fix $0<c<M$, take $\tau=1/M$, and let $h\equiv(c+M)/(2M)$. Then $h\in[c\tau,M\tau)$~and
\[
\Cov_{\mathrm{mix}}(\tau)=\Pr[h\ge1]=0,
\qquad
\Cov_{\mathrm{orac}}(c\tau)=1.
\]
This construction violates the proposed bound for every $0<c<M$, proving that $M$ is optimal.~\qed

\section{Training, estimation, and diagnostics}
\label{app:protocol}

This appendix specifies the training, rate estimators, matched-budget comparisons, and router inputs needed to interpret the deployment results. It also checks whether the findings depend on the reliability band or evaluated~\mbox{split}.

Comparisons use the same held-out prompts. The primary deployment comparisons match response counts. The pool-size diagnostics use rates estimated from all available responses. Routers are fitted using validation outcomes and predict from prompt features, with no access to test outcomes. Text routers use text alone, while embedding routers also use the image. The plug-in oracle is an analysis tool that uses estimated rates on the scored~\mbox{prompts}.

\subsection{Training and rate estimation}

\paragraph{How similar are the trained policies?} Similar average accuracy can conceal different strengths across prompts. Pairwise correlations of the members' estimated per-prompt rates average $0.938$, compared with $0.944$ under the re-deal null. Rates differ by more than $0.25$ on $21\%$ of prompts, compared with a null mean of $18\%$ and range of $14$--$22\%$. These diagnostics show similar estimated difficulty profiles, providing context for the small excesses in the oracle-gap~\mbox{tests}.

\paragraph{Training configuration.} File hashes confirm that the five runs use the same base checkpoint, ordered list of $400$ training prompts, and configuration apart from random-seed fields. The seeds control new-parameter initialization, training-data sampling, and response generation. Each run trains rank-$16$ low-rank adaptation (LoRA) modules on the language model's attention and feed-forward projections. The image-processing network remains frozen. Training uses GRPO for $500$ steps with learning rate $10^{-5}$, eight responses per prompt, and no Kullback--Leibler penalty constraining departure from a reference policy. The stronger recipe of Section~\ref{sec:specialist} keeps this configuration and changes three settings: $1{,}500$ steps, the full training list ($1{,}200$ Geometry3K prompts or all $366$ MathVista training prompts, disjoint from validation and test), and rank-$64$ adapters with scaling $128$, with two seeds per~\mbox{dataset}.

\paragraph{Evaluation sampling.} Each policy generates $16$ responses per prompt at sampling temperature $1$, in one batch per prompt. Each policy is evaluated in a separate task with the same initial random seed. Different response lengths cause the random-number streams to diverge from the first prompt onward. Appendix~\ref{app:audit} checks whether responses with matching sample indices show excess agreement across policies. Geometry3K uses the public release of \citet{lu2021intergps}. MathVista uses the public \emph{testmini} split~\citep{lu2024mathvista}.

\paragraph{What $16$ responses can establish.} A small response sample cannot identify low-threshold coverage exactly. Even with zero successes in $16$ responses, the one-sided $95\%$ Clopper--Pearson upper bound on the true success probability is $0.171$ \citep{clopper1934}. A zero count therefore does not certify that a policy falls below the lower reliability threshold. All empirical coverage decisions should be read as estimates at the stated~\mbox{budget}.

\paragraph{Raw and smoothed rate estimates.} Alongside the raw frequency, we use a beta-binomial model to smooth estimates from $k=16$ responses. Let $s$ be the correct-response count for one policy and prompt, and assign its unknown success probability a $\mathrm{Beta}(\alpha,\beta)$ prior. The posterior mean~is
\[
\hat p=\frac{s+\alpha}{k+\alpha+\beta}.
\]
The positive parameters $\alpha$ and $\beta$ are fitted by matching moments across prompt--policy pairs within each evaluated split, including the base model. They are $(0.347,0.601)$ on Geometry3K test, $(0.323,0.606)$ on Geometry3K validation, $(0.453,0.272)$ on MathVista test, and $(0.497,0.234)$ on MathVista validation. The posterior mean averages the raw rate $s/k$ with the prior mean $\alpha/(\alpha+\beta)$, giving the prior weight~$(\alpha+\beta)/(k+\alpha+\beta)$.

Although the prior is fitted on the evaluated split, smoothing changes no single-member decision at the primary reporting thresholds. Raw and posterior rates therefore give the same primary $D$ and $L$. They can give different mixture coverage because averaging members produces a finer rate grid. All re-deal comparisons use raw frequencies for both observed and randomized outcomes. At $k=64$ on Geometry3K, smoothing changes the count required at $\tau=0.05$ from four successes to three. It changes $D$ from $0.133$ to $0.167$ and $L$ from $0.050$ to $0.083$. The other reported Geometry3K thresholds and all reported MathVista thresholds give identical raw and smoothed $D$ and $L$. Prompt-bootstrap intervals keep the fitted prior parameters~\mbox{fixed}.

\paragraph{Matching the mixture's response budget.} Averaging all five members' rates uses $80$ responses per prompt, compared with $16$ for a single member. This gives the mixture a finer raw count grid. At $\tau=0.10$, the pooled estimate requires $8$ successes out of $80$. A member requires $2$ out of $16$, or $0.125$. Its rate estimate also has lower sampling variance under independent draws. Because thresholded estimates depend on both the grid and the variance, the primary comparison subsamples $16$ of the $80$ outcomes per prompt. It applies the split's fitted prior exactly as for one member and averages rAUC over $200$~\mbox{subsamples}.

On Geometry3K test, this procedure gives mixture rAUC $0.592$, compared with $0.597$ when the posterior rate is estimated from all $80$ pooled outcomes and $0.607$ for the best member. The corresponding MathVista values are $0.534$, $0.536$, and $0.543$. Intervals use paired prompt bootstraps over the same subsamples. Matching response count aligns the count grids, but subsampling a fixed pool is still an evaluation approximation to fresh mixture~\mbox{sampling}.

\subsection{Comparator and routing diagnostics}

\paragraph{Weight averaging.} Each LoRA update has the form $sB_iA_i$, where $A_i$ and $B_i$ are the learned low-rank factors for seed $i$ and $s$ is the adapter scaling constant. The model soup uses the mean update $\frac15\sum_i sB_iA_i$. We store this sum as one rank-$80$ adapter by stacking the five factor pairs, allowing evaluation through the same code path as a seed policy. The largest numerical deviation from the explicit mean update is~$5\times10^{-11}$.

At $k=16$, paired-bootstrap intervals for the soup's rAUC difference from the best member are $[-0.046,+0.014]$ on Geometry3K test and $[-0.031,+0.019]$ on MathVista test. Against the budget-matched mixture, they are $[-0.021,+0.028]$ and $[-0.023,+0.019]$. These comparisons support the main table's finding that parameter averaging yields no resolved~\mbox{advantage}.

\paragraph{Comparator selection.} For rAUC summaries, the best member maximizes rAUC on the evaluated split unless otherwise stated. This retrospective selection requires access to test outcomes. We therefore also report comparisons with the member selected on validation. Coverage oracle gaps $L(\tau)$ use the best coverage at each threshold, as defined in Section~\ref{sec:setup}. A router's break-even ratio must use the same threshold-specific comparator as its reported coverage~\mbox{gain}.

\paragraph{Text-based routers.} All routers use validation outcomes for training and have no access to test labels. The linear text router fits multinomial logistic regression to predict the member with the highest validation rate from individual words and adjacent word pairs. The gradient-boosted router fits one decision-tree ensemble per member to predict its rate from features such as prompt length, numeral counts, typesetting markers, and geometry keywords. It selects the member with the highest predicted rate. We report the median over initializations. Coverage varies by at most $0.017$ on Geometry3K and $0.005$ on~\mbox{MathVista}.

The text nearest-neighbor router represents each prompt with term frequency--inverse document frequency (TF-IDF) features, which weight words by their frequency in the prompt and rarity across prompts. It finds the $15$ validation prompts with the largest cosine similarity and selects the member with the highest mean success rate on those neighbors. This tests whether local text similarity predicts member performance without fitting a~\mbox{classifier}.

\paragraph{Routers that also use the image.} The two embedding families use the base model's final hidden representations before any response is generated. Averaging these representations over all image and text tokens gives a $3584$-dimensional vector for each prompt. The linear router standardizes these vectors, reduces them to $64$ principal components, and fits multinomial logistic regression. We report the median of three initializations. The embedding nearest-neighbor router uses cosine similarity and $15$ neighbors. These families test whether including the diagram changes the result obtained with text alone. Table~\ref{tab:routing} reports every family and threshold. No gain interval lies wholly above~\mbox{zero}.

\begin{table}[p]
\centering
\caption{\textbf{Prompt-based routing at each reliability threshold.} Routers are fitted on validation. Embedding families use image and text representations from the base model. \emph{Gain} compares coverage with the threshold-specific best member on the evaluated split. \emph{Gain (val.)} uses the validation-selected member. Intervals are $95\%$ prompt-bootstrap intervals. The disagreement error $e_D$ counts selections below threshold on disagreement prompts, and $L/D$ is its break-even~value.}
\label{tab:routing}
\scriptsize
\renewcommand{\arraystretch}{1.0}
\setlength{\tabcolsep}{5pt}
\begin{tabular}{@{}l l S[table-format=+1.3] c S[table-format=+1.3] S[table-format=1.3] S[table-format=1.3]@{}}
\toprule
\textbf{$\tau$} & \textbf{Family} & {\bfseries gain} & \textbf{$95\%$ interval} & {\bfseries gain (val.)} & {\bfseries $e_D$} & {\bfseries $L/D$} \\
\midrule
\multicolumn{7}{@{}l}{\emph{Geometry3K test split, $240$ prompts}} \\
\midrule
$0.05$ & linear            & -0.008 & $[-0.062,+0.008]$ & -0.008 & 0.507 & 0.479 \\
$0.05$ & gradient-boosted  & +0.004 & $[-0.046,+0.021]$ & +0.004 & 0.465 & 0.479 \\
$0.05$ & nearest-neighbor  & +0.004 & $[-0.042,+0.025]$ & +0.004 & 0.465 & 0.479 \\
$0.05$ & embedding, linear  & -0.017 & $[-0.054,+0.021]$ & -0.017 & 0.535 & 0.479 \\
$0.05$ & embedding, nearest-neighbor & -0.025 & $[-0.058,+0.008]$ & -0.025 & 0.563 & 0.479 \\
\midrule
$0.10$ & linear            & -0.008 & $[-0.054,+0.000]$ & +0.000 & 0.518 & 0.482 \\
$0.10$ & gradient-boosted  & -0.004 & $[-0.050,+0.013]$ & +0.004 & 0.500 & 0.482 \\
$0.10$ & nearest-neighbor  & +0.017 & $[-0.033,+0.038]$ & +0.025 & 0.411 & 0.482 \\
$0.10$ & embedding, linear  & -0.021 & $[-0.058,+0.017]$ & -0.012 & 0.571 & 0.482 \\
$0.10$ & embedding, nearest-neighbor & +0.000 & $[-0.037,+0.037]$ & +0.008 & 0.482 & 0.482 \\
\midrule
$0.20$ & linear            & -0.008 & $[-0.025,+0.000]$ & +0.021 & 0.400 & 0.360 \\
$0.20$ & gradient-boosted  & -0.046 & $[-0.083,-0.012]$ & -0.017 & 0.580 & 0.360 \\
$0.20$ & nearest-neighbor  & -0.050 & $[-0.092,-0.017]$ & -0.021 & 0.600 & 0.360 \\
$0.20$ & embedding, linear  & -0.021 & $[-0.050,+0.004]$ & +0.008 & 0.460 & 0.360 \\
$0.20$ & embedding, nearest-neighbor & -0.033 & $[-0.063,-0.004]$ & -0.004 & 0.520 & 0.360 \\
\midrule
$0.30$ & linear            & -0.013 & $[-0.042,+0.000]$ & +0.021 & 0.447 & 0.383 \\
$0.30$ & gradient-boosted  & -0.033 & $[-0.075,-0.004]$ & +0.000 & 0.553 & 0.383 \\
$0.30$ & nearest-neighbor  & -0.037 & $[-0.079,-0.017]$ & -0.004 & 0.574 & 0.383 \\
$0.30$ & embedding, linear  & -0.008 & $[-0.042,+0.021]$ & +0.025 & 0.426 & 0.383 \\
$0.30$ & embedding, nearest-neighbor & -0.037 & $[-0.075,+0.000]$ & -0.004 & 0.574 & 0.383 \\
\midrule
\multicolumn{7}{@{}l}{\emph{MathVista test split, $184$ prompts}} \\
\midrule
$0.5$ & linear            & -0.016 & $[-0.049,+0.000]$ & +0.000 & 0.500 & 0.350 \\
$0.5$ & gradient-boosted  & -0.005 & $[-0.038,+0.016]$ & +0.011 & 0.400 & 0.350 \\
$0.5$ & nearest-neighbor  & -0.027 & $[-0.060,-0.005]$ & -0.011 & 0.600 & 0.350 \\
$0.5$ & embedding, linear  & -0.016 & $[-0.049,+0.016]$ & +0.000 & 0.500 & 0.350 \\
$0.5$ & embedding, nearest-neighbor & -0.011 & $[-0.043,+0.022]$ & +0.005 & 0.450 & 0.350 \\
\midrule
$0.6$ & linear            & -0.027 & $[-0.060,+0.000]$ & +0.000 & 0.524 & 0.286 \\
$0.6$ & gradient-boosted  & -0.016 & $[-0.049,+0.005]$ & +0.011 & 0.429 & 0.286 \\
$0.6$ & nearest-neighbor  & -0.033 & $[-0.065,-0.011]$ & -0.005 & 0.571 & 0.286 \\
$0.6$ & embedding, linear  & -0.033 & $[-0.060,-0.011]$ & -0.005 & 0.571 & 0.286 \\
$0.6$ & embedding, nearest-neighbor & -0.033 & $[-0.065,-0.005]$ & -0.005 & 0.571 & 0.286 \\
\midrule
$0.7$ & linear            & -0.016 & $[-0.065,+0.000]$ & -0.011 & 0.484 & 0.387 \\
$0.7$ & gradient-boosted  & -0.022 & $[-0.060,+0.000]$ & -0.016 & 0.516 & 0.387 \\
$0.7$ & nearest-neighbor  & -0.027 & $[-0.071,-0.005]$ & -0.022 & 0.548 & 0.387 \\
$0.7$ & embedding, linear  & -0.016 & $[-0.060,+0.027]$ & -0.011 & 0.484 & 0.387 \\
$0.7$ & embedding, nearest-neighbor & -0.016 & $[-0.054,+0.016]$ & -0.011 & 0.484 & 0.387 \\
\midrule
$0.8$ & linear            & -0.043 & $[-0.092,+0.000]$ & -0.022 & 0.600 & 0.371 \\
$0.8$ & gradient-boosted  & -0.027 & $[-0.065,+0.000]$ & -0.005 & 0.514 & 0.371 \\
$0.8$ & nearest-neighbor  & -0.016 & $[-0.060,+0.011]$ & +0.005 & 0.457 & 0.371 \\
$0.8$ & embedding, linear  & -0.027 & $[-0.071,+0.016]$ & -0.005 & 0.514 & 0.371 \\
$0.8$ & embedding, nearest-neighbor & -0.011 & $[-0.043,+0.022]$ & +0.011 & 0.429 & 0.371 \\
\midrule
$0.9$ & linear            & -0.049 & $[-0.092,-0.016]$ & -0.027 & 0.647 & 0.382 \\
$0.9$ & gradient-boosted  & -0.016 & $[-0.054,+0.016]$ & +0.005 & 0.471 & 0.382 \\
$0.9$ & nearest-neighbor  & -0.022 & $[-0.049,-0.005]$ & +0.000 & 0.500 & 0.382 \\
$0.9$ & embedding, linear  & -0.016 & $[-0.054,+0.016]$ & +0.005 & 0.471 & 0.382 \\
$0.9$ & embedding, nearest-neighbor & -0.033 & $[-0.071,+0.005]$ & -0.011 & 0.559 & 0.382 \\
\bottomrule
\end{tabular}
\end{table}

\subsection{Sensitivity and uncertainty}

\paragraph{Reliability-band sensitivity.} To check robustness to the reliability band, we evaluate the budget-matched mixture on $[0.05,0.30]$, $[0.05,0.50]$, $[0.10,0.30]$, and $[0.05,0.20]$. In that order, its rAUC differences from the test-selected best member are $(-0.017,\allowbreak -0.013,\allowbreak -0.022,\allowbreak -0.009)$. The interval endpoints span $-0.038$ to $+0.012$. Differences from the validation-selected member are $(-0.002,\allowbreak -0.001,\allowbreak -0.002,\allowbreak -0.001)$, with every interval including zero. No band gives a detectable improvement. The primary-band estimate differs slightly from Section~\ref{sec:protocol} because this analysis uses an independent set of random~\mbox{subsamples}.

\paragraph{Uncertainty and computational budgets.} The prompt is the resampling unit. Each paired-bootstrap replicate samples prompts with replacement and uses the same sampled prompts for both comparison arms. Router gain intervals in Table~\ref{tab:routing} use $2{,}000$ replicates and re-select the best member within each replicate, which shifts them against the router relative to the point estimate, whose best member is fixed on the full split. The cross-dataset pools hold the comparator fixed. These intervals measure prompt-sampling uncertainty for the five fixed policies. They exclude variability from retraining. The $26$ overlapping pool subsets share members, so they cannot serve as independent replications. Permutation $p$-values are the fractions of randomized statistics at least as large as the observed value. Individual values are unadjusted unless stated otherwise. Section~\ref{sec:protocol} reports the Benjamini--Hochberg adjustment across $156$ tests. Mean completion lengths differ by up to $12\%$ on Geometry3K and $8\%$ on MathVista, giving different token costs even at matched response~\mbox{counts}.

\section{Null models, power, and sampling bias}
\label{app:audit}

This appendix defines the nulls and evaluates the two questions on which their interpretation depends: which departures they detect, and whether those departures are useful for routing. Simulations also test whether subtracting a null mean estimates the population gap. Complete split-level results follow these~\mbox{checks}.

\subsection{Null models and independent scoring}

\paragraph{Why the plug-in gap can be biased.} The oracle selects the largest estimated rate on each prompt, favoring positive sampling errors \citep{fowler2026capfrontier,chen2026routinggap}. The best-member comparator is also selected using noisy estimates. This second selection can inflate the coverage subtracted from the oracle, acting in the opposite direction. Thresholding introduces additional sensitivity to rate-estimation error. Consequently, the plug-in gap is not automatically an upper bound on the population gap. We compare it with the null distributions and also score selected members on separate~\mbox{responses}.

\paragraph{Why conditioning removes the unknown parameters.} Under the additive-logit model in Section~\ref{sec:protocol}, the joint probability of a count table~is
\[
\Pr(c\mid a,b)=
\left[\prod_{i,m}\binom{k}{c_{im}}\right]
\frac{\exp\!\left(\sum_i a_i\sum_m c_{im}+\sum_m b_m\sum_i c_{im}\right)}
{\prod_{i,m}(1+e^{a_i+b_m})^k}.
\]
Once prompt and member totals are fixed, the fraction is constant across feasible tables. Their conditional probabilities therefore depend only on the binomial weights. This is the conditional no-interaction test for an additive-logit, or Rasch, model, related to differential-item-functioning tests in item response theory \citep{rasch1960probabilistic,holland1988dif}. The plain re-deal null sets $b_m=0$ for every member and conditions only on prompt totals. Rejection of the margin-preserving model indicates a departure from additivity on the log-odds scale. Identifying useful specialists then requires examining which members succeed and whether prompt features predict that~\mbox{success}.

\paragraph{Sampling the margin-preserving null.} We sample count tables with the conditional weights above. A Metropolis Markov chain proposes a $2\times2$ transfer: move one success from member $a$ to member $b$ on one prompt, and from $b$ to $a$ on another. This preserves both prompt and member totals. Feasible proposals are accepted according to the ratio of the target table weights. The chain discards its first $60nM$ proposals, then retains every $2nM$th table until $2{,}000$ samples have been collected. Two independent chains agree on the null mean within $0.001$ and on $p_{\mathrm{m}}$ within $0.011$. The reported test approximates the exact conditional law using these sampled tables. Sampling feasible tables uniformly would give a different null, because the binomial weights account for the number of outcome assignments represented by each~\mbox{table}.

Both nulls preserve the total successes on every prompt. An effect shared by all members on a prompt, including a shared response to a verifier defect, is therefore retained. Member-specific effects can change under randomization. Primary empirical tests use $2{,}000$ null samples. The simulation studies below specify their smaller permutation budgets. On Geometry3K test at $\tau=0.10$, the margin-preserving mean is $0.095$, compared with $0.096$ for the plain null, and~$p_{\mathrm{m}}=0.063$.

\paragraph{Split-half selection and scoring.} For each prompt, we choose a member from eight responses per policy and score it on the remaining eight. We average over $200$ random splits. With eight responses, both $\tau=0.05$ and $0.10$ require one success, so their empirical coverage decisions use the same count threshold. The eight-response plug-in gaps are $0.146$ and $0.148$ at $\tau=0.05$ and $0.10$. When the best fixed comparator is also chosen on the selection half, the split-half gaps are $(+0.021,\allowbreak +0.020,\allowbreak +0.001,\allowbreak -0.002)$ at the four ascending thresholds. Choosing the comparator on the scoring half instead gives $(-0.006,\allowbreak -0.007,\allowbreak -0.015,\allowbreak -0.017)$. That comparison favors the fixed member by letting it benefit from selection on the same responses used to score~it.

\subsection{Calibration and power}

\paragraph{Concentrated specialization.} We transfer successes within each prompt from other members to a randomly chosen specialist, keeping the prompt total fixed. Each member is equally likely to be designated as the specialist. Its total successes can change through these transfers. A perturbation strength $\delta$ sets the target to $\operatorname{round}(\delta k)$ transfers per prompt. A transfer moves one success from a randomly chosen donor with a positive count to the specialist. Transfers stop if the specialist reaches $k$ successes or no donor has a success left. Table~\ref{tab:power} reports target counts, so some prompts receive fewer transfers. We rerun both tests at $\tau=0.10$ for $20$ trials per strength, using $400$ null samples per trial. The margin-preserving chain in these power trials discards $30nM$ proposals and then retains one table every $nM$~\mbox{proposals}.

At $\delta=0$, the counts remain unchanged. Variation in rejection comes from finite Monte Carlo sampling of the null. With a target of one transfer, the plain and margin-preserving tests reject at level $0.05$ in $65\%$ and $70\%$ of plantings. Both reject in every planting with a target of two transfers. The mean excess with one transfer is $+0.024$, compared with the observed $+0.017$. This comparison measures sensitivity to the specified transfers. The relationship between excess and population gap is examined~\mbox{below}.

\begin{table}[!ht]
\centering
\caption{\textbf{Sensitivity to concentrated specialization on Geometry3K test.} The perturbation strength $\delta$ sets the transfer count reported as \emph{target}, the maximum integer count. \emph{Power} is the reported rejection fraction across $20$ trials at level $0.05$. \emph{Excess} is the planted $L(0.10)$ minus the plain-null mean. The $\delta=0$ row repeats the tests on the unchanged observed pool. False-positive rates are assessed separately using null-generated~datasets.}
\label{tab:power}
\small
\begin{tabular}{@{}S[table-format=1.2] S[table-format=1] S[table-format=1.2] S[table-format=1.2] S[table-format=+1.3]@{}}
\toprule
{$\delta$} & {target} & {power, re-deal} & {power, margin} & {excess} \\
\midrule
0.00 & 0 & 0.10 & 0.15 & +0.017 \\
0.05 & 1 & 0.65 & 0.70 & +0.024 \\
0.10 & 2 & 1.00 & 1.00 & +0.049 \\
0.15 & 2 & 1.00 & 1.00 & +0.052 \\
0.20 & 3 & 1.00 & 1.00 & +0.070 \\
0.25 & 4 & 1.00 & 1.00 & +0.076 \\
\bottomrule
\end{tabular}
\end{table}

\paragraph{False-positive rates under the nulls.} We generate $200$ datasets under each null from the Geometry3K test outcomes. Plain-null datasets come from independent re-deals within prompts. Margin-preserving datasets come from retained tables in the conditional chain. We then rerun both tests with $400$ permutations per dataset. On plain-null datasets, the plain test rejects in $4.0\%$ and $3.0\%$ of cases at $\tau=0.05$ and $0.10$, while the margin-preserving test rejects in $3.5\%$ and $3.0\%$. On margin-preserving datasets, the corresponding rates are $5.5\%$, $3.0\%$, $6.0\%$, and $3.0\%$. These simulations are consistent with the nominal $5\%$ level, with somewhat conservative behavior at~$\tau=0.10$.

\paragraph{Diffuse rate differences.} Concentrating transfers on one specialist is only one alternative to the null. We also perturb every member's log-odds of success on every prompt by an independent mean-zero Gaussian term with standard deviation $\sigma$, centered on the prompt's observed mean rate. Before taking log-odds, we clip that mean to $[\epsilon,1-\epsilon]$, with $\epsilon=1/(2Mk)$ and $k=16$ the observed response count. No member is designated as the sole specialist. Both tests are repeated on $20$ simulated datasets per $\sigma$, with $400$ permutations each (Table~\ref{tab:diffuse}). At $\sigma=0.5$, the population gap is about $0.05$ and both tests have power $0.50$. At $\sigma=1.0$, the gap is about $0.11$ and both have power $1.00$. At $\sigma=0.25$, a gap of about $0.02$ is rarely detected. Increasing the response budget to $64$ raises power to $0.90$ at a gap of $0.05$ and $0.15$ at a gap of~$0.02$.

\subsection{What the excess measures}

We simulate common prompt rates from the Beta distribution fitted to Geometry3K test ($\alpha=0.347$, $\beta=0.601$). Each prompt independently receives a specialist perturbation with probability $f$. On a selected prompt with common rate $q$, one uniformly chosen member receives rate $\min(1,q+s)$. The others retain rate $q$. We sample $k=16$ responses per member, then compare the known population gap $L(0.10)$ with the plug-in gap $\hat L$ and its excess over the plain-null mean. Each setting uses $30$ replicates and $300$ re-deals per replicate (Table~\ref{tab:sims}). For population gaps of at least $0.02$, the excess recovers only $14\%$--$35\%$ of the gap. Ratios are unstable for smaller gaps. With no specialists, the mean excess is $+0.001$. At $k=64$, the recovered share rises to~$44\%$--$71\%$.

\begin{table}[!ht]
\centering
\caption{\textbf{Finite-sample excess versus the population oracle gap.} Simulated prompt rates use the fitted Geometry3K Beta distribution. A fraction $f$ of prompts receives a specialist perturbation of strength $s$, followed by $k=16$ sampled responses per member. Entries average $30$ replicates. $L$ is the known population gap at $\tau=0.10$, and $\hat L$ is the plug-in gap. \emph{Excess} subtracts the plain-null mean from $\hat L$, and \emph{ratio} is the excess divided by the population~gap.}
\label{tab:sims}
\small
\setlength{\tabcolsep}{6pt}
\begin{tabular}{@{}S[table-format=1.1] S[table-format=1.2] S[table-format=1.3] S[table-format=1.3] S[table-format=+1.3] S[table-format=1.2]@{}}
\toprule
{$f$} & {$s$} & {$L$} & {$\hat L$} & {excess} & {ratio} \\
\midrule
0.1 & 0.05 & 0.003 & 0.088 & +0.000 & 0.01 \\
0.1 & 0.10 & 0.021 & 0.091 & +0.003 & 0.14 \\
0.1 & 0.20 & 0.023 & 0.098 & +0.004 & 0.19 \\
0.3 & 0.05 & 0.012 & 0.099 & +0.006 & 0.53 \\
0.3 & 0.10 & 0.070 & 0.113 & +0.013 & 0.19 \\
0.3 & 0.20 & 0.076 & 0.135 & +0.024 & 0.32 \\
0.5 & 0.05 & 0.027 & 0.104 & +0.009 & 0.33 \\
0.5 & 0.10 & 0.124 & 0.132 & +0.028 & 0.22 \\
0.5 & 0.20 & 0.128 & 0.167 & +0.044 & 0.35 \\
\bottomrule
\end{tabular}
\end{table}

\subsection{Positive controls and model sensitivity}

\paragraph{Combining two model sizes.} We join the five 7B policies with the five 3B policies on their shared $240$ Geometry3K test prompts. The two groups have mean success rates $0.369$ and $0.240$, so this control includes a substantial quality difference. At thresholds $(0.05,\allowbreak 0.10,\allowbreak 0.20,\allowbreak 0.30)$, the observed oracle gaps are $(0.158,\allowbreak 0.150,\allowbreak 0.121,\allowbreak 0.104)$. The plain-null means are $(0.149,\allowbreak 0.148,\allowbreak 0.133,\allowbreak 0.127)$, with upper-tail fractions $(0.19,\allowbreak 0.53,\allowbreak 0.88,\allowbreak 0.98)$ (Table~\ref{tab:master}). Redistributing the stronger group's successes to weaker members can itself create a large estimated~gap.

The margin-preserving null retains the group quality difference. Its means are $(0.115,\allowbreak 0.103,\allowbreak 0.092,\allowbreak 0.085)$, with $p_{\mathrm{m}}=(<0.001,\allowbreak <0.001,\allowbreak 0.010,\allowbreak 0.057)$. Thus the low-threshold gaps exceed what this additive-logit model predicts. We examine sensitivity to the link function next. The 3B pool alone resembles the primary 7B pool. Its test gap at $\tau=0.10$ is $0.121$, below both null means~(Table~\ref{tab:master}).

\paragraph{Sensitivity to the link function.} The margin-preserving null defines a constant quality effect on the log-odds scale. A constant effect on another scale can violate this assumption without creating the intended form of specialization. We simulate two groups of five members using the observed Geometry3K prompt difficulties. Group means match either the $0.37$ versus $0.24$ quality difference in the mixed-size pool or a larger $0.37$ versus $0.17$ difference. We impose constant member effects on three scales: logit, probit (the inverse standard-normal distribution function), and probability itself~(Table~\ref{tab:link}).

At $\tau\le0.10$, margin-preserving rejection rates are at most $0.13$ across these settings. The plain test never rejects. At $\tau=0.20$, the additive-probability setting rejects in $0.33$ of replicates, making the mixed-size rejection there sensitive to model misspecification. At $\tau=0.10$, the simulated plug-in gaps from quality differences alone are about $0.09$--$0.10$, compared with $0.150$ in the observed mixed-size~\mbox{pool}.

\begin{table}[!ht]
\centering
\caption{\textbf{Quality differences that are constant on different scales.} Two groups of five members share Geometry3K prompt difficulties. The stronger group has mean rate $0.37$. The table gives the weaker group's mean and the \emph{gap} between them. Each row uses $k=16$, $30$ replicates, and $400$ permutations per test. \emph{Reject} is the margin-preserving rejection fraction at nominal level $0.05$. For non-logit rows, rejection can reflect misspecification of the null's link. The plain re-deal never rejects in these simulations. $L$ is the mean plug-in oracle~gap.}
\label{tab:link}
\scriptsize
\setlength{\tabcolsep}{4pt}
\begin{tabular}{@{}l S[table-format=1.2] S[table-format=1.2] S[table-format=1.2] S[table-format=1.2] S[table-format=1.2] S[table-format=1.3] S[table-format=1.3] S[table-format=1.3]@{}}
\toprule
& & & \multicolumn{3}{c}{reject, margin-preserving} & \multicolumn{3}{c}{$L$} \\
\cmidrule(lr){4-6}\cmidrule(lr){7-9}
\textbf{Link} & {gap} & {weak mean} & {$\tau{=}0.05$} & {$0.10$} & {$0.20$} & {$0.05$} & {$0.10$} & {$0.20$} \\
\midrule
logit (the null's model) & 0.13 & 0.24 & 0.03 & 0.13 & 0.03 & 0.140 & 0.103 & 0.079 \\
 & 0.20 & 0.17 & 0.03 & 0.00 & 0.07 & 0.133 & 0.095 & 0.083 \\
probit & 0.13 & 0.24 & 0.00 & 0.07 & 0.10 & 0.135 & 0.098 & 0.083 \\
 & 0.20 & 0.17 & 0.00 & 0.00 & 0.00 & 0.130 & 0.091 & 0.080 \\
additive probability & 0.13 & 0.24 & 0.00 & 0.00 & 0.33 & 0.124 & 0.096 & 0.077 \\
 & 0.20 & 0.17 & 0.00 & 0.00 & 0.00 & 0.123 & 0.094 & 0.080 \\
\bottomrule
\end{tabular}
\end{table}

\paragraph{Can routers exploit the positive controls?} We fit the embedding routers on validation outcomes and score them on the corresponding test pools (Table~\ref{tab:positive}). On the 7B+3B pool at $\tau=0.10$, the nearest-neighbor and linear routers have disagreement errors $0.36$ and $0.43$, compared with break-even $0.36$. Their gains are $0.000$ and $-0.029$. The validation-selected 7B member already matches the best member at this threshold. These routers therefore do not recover a gain from the departure detected by the margin-preserving~\mbox{test}.

For predictable specialization, we fit the embedding standardization, first principal component, and quintile boundaries on validation prompts. Each of the five resulting intervals designates one specialist. We apply this fixed rule to test prompts and transfer up to the target number of successes, subject to the same availability constraints as above. With a target of two transfers per prompt, both tests reject at $\tau=0.10$. The nearest-neighbor router gains $+0.079$ of the $0.146$ oracle gap, with disagreement error $0.21$. A selector given the specialist assignment recovers the full gap at this threshold. With a target of one transfer, the nearest-neighbor gain is $+0.037$, with an interval touching zero. These controls show that a detected difference can yield a routing gain when the specialist assignment is predictable from the router's~\mbox{features}.

\begin{table}[!ht]
\centering
\caption{\textbf{Routing on positive controls.} The mixed-size pool contains five 7B and five 3B policies evaluated on shared Geometry3K prompts. The planted pools use five 7B policies with the specialist assigned by the quintile of the first principal component of the prompt embedding. All statistics shown are test estimates. Each router block gives coverage gain over the threshold-specific best member and disagreement error $e_D$. The break-even error is $L/D$. Routers are trained on validation with the same planting rule. The final block uses the validation-selected 7B member for the mixed-size pool and the known specialist assignment for planted~pools.}
\label{tab:positive}
\scriptsize
\setlength{\tabcolsep}{3pt}
\begin{tabular}{@{}l S[table-format=1.3] S[table-format=1.3] S[table-format=1.3] S[table-format=1.3] S[table-format=1.2] S[table-format=+1.3] S[table-format=1.2] S[table-format=+1.3] S[table-format=1.2] S[table-format=+1.3] S[table-format=1.2]@{}}
\toprule
& & & & & & \multicolumn{2}{c}{nearest-neighbor} & \multicolumn{2}{c}{linear} & \multicolumn{2}{c}{size rule / specialist} \\
\cmidrule(lr){7-8}\cmidrule(lr){9-10}\cmidrule(lr){11-12}
{pool} & {$L$} & {$D$} & {$p$} & {$p_{\mathrm{m}}$} & {$L/D$} & {gain} & {$e_D$} & {gain} & {$e_D$} & {gain} & {$e_D$} \\
\midrule
7B+3B, $\tau=0.05$ & 0.158 & 0.442 & 0.223 & {$<$0.001} & 0.36 & -0.029 & 0.42 & -0.029 & 0.42 & +0.000 & 0.36 \\
7B+3B, $\tau=0.10$ & 0.150 & 0.421 & 0.539 & {$<$0.001} & 0.36 & +0.000 & 0.36 & -0.029 & 0.43 & +0.000 & 0.36 \\
7B+3B, $\tau=0.20$ & 0.121 & 0.454 & 0.886 & 0.009 & 0.27 & -0.033 & 0.34 & -0.042 & 0.36 & -0.029 & 0.33 \\
\midrule
planted $1$ of $16$, $\tau=0.05$ & 0.150 & 0.317 & 0.002 & {$<$0.001} & 0.47 & +0.075 & 0.24 & +0.021 & 0.41 & +0.150 & 0.00 \\
planted $1$ of $16$, $\tau=0.10$ & 0.121 & 0.263 & 0.011 & 0.008 & 0.46 & +0.037 & 0.32 & +0.008 & 0.43 & +0.075 & 0.17 \\
planted $2$ of $16$, $\tau=0.05$ & 0.171 & 0.333 & {$<$0.001} & {$<$0.001} & 0.51 & +0.100 & 0.21 & +0.046 & 0.38 & +0.171 & 0.00 \\
planted $2$ of $16$, $\tau=0.10$ & 0.146 & 0.317 & {$<$0.001} & {$<$0.001} & 0.46 & +0.079 & 0.21 & +0.029 & 0.37 & +0.146 & 0.00 \\
planted $2$ of $16$, $\tau=0.20$ & 0.117 & 0.267 & {$<$0.001} & {$<$0.001} & 0.44 & +0.054 & 0.23 & +0.004 & 0.42 & +0.100 & 0.06 \\
\bottomrule
\end{tabular}
\end{table}

\subsection{Complete results and additional diagnostics}

\paragraph{Validation and MathVista results.} No oracle gap on Geometry3K validation or either MathVista split significantly exceeds the plain re-deal null at level $0.05$ (Table~\ref{tab:master}). Of these $14$ conditions, $12$ lie within the null's $90\%$ range. Geometry3K validation at $\tau=0.10$ and MathVista validation at $\tau=0.90$ lie below its fifth percentile. A broad advantage for one member can produce a small oracle gap by raising best-member coverage more than oracle coverage. We examine the Geometry3K case~\mbox{below}.

\paragraph{Why one validation gap falls below both nulls.} On Geometry3K validation, seed $5$ leads, and the observed gap at $\tau=0.10$ is $0.067$, below plain and margin-preserving means of $0.105$ and $0.103$. Because generation starts from the same random seed for each policy, we check whether matching response indices have unusually similar outcomes. Correctness agreement is $0.779$ for matching indices and $0.781$ after shuffling. Response-length correlations are also similar. The following rate pattern is consistent with the low~gap.

Seed $5$ covers $0.679$ of validation prompts at $\tau=0.10$, compared with conditional mean $0.647$ given its total successes. The lower-tail fraction is $0.993$. Its successes therefore extend across more prompts than the additive model typically predicts. A broad, shallow advantage for the leading seed leaves fewer prompts covered only by another member. This departure from the null reduces the oracle~gap.

\paragraph{Lower tails and prompts with no randomization variance.} Table~\ref{tab:master} reports both upper- and lower-tail fractions. The lower tail asks whether an observed gap is unusually small under a null. Among the $43$ single-size conditions, the plain-null lower tail is below $0.05$ on Geometry3K validation at $\tau=0.10$ and MathVista validation at $\tau=0.90$. These checks span dependent conditions. Between $17$ and $90$ prompts per set have either no successes or no failures across the entire pool. Such \emph{degenerate} prompts contribute no permutation variance. They comprise $7\%$--$31\%$ of each set, reducing the number of prompts that can inform the~\mbox{test}.

\paragraph{An oracle for average accuracy.} To check whether the result depends on thresholding, we also test a $\mathrm{Pass@1}$ oracle gap. Its statistic is the average over prompts of the largest estimated member rate, minus the largest member's average rate. On Geometry3K test, the gap with raw rates is $0.079$ against plain-null mean $0.078$ ($p=0.43$, $p_{\mathrm{m}}=0.21$). Validation gives $0.074$ against $0.077$ ($p=0.75$). On MathVista test, the gap is $0.069$ against $0.062$ ($p=0.044$, $p_{\mathrm{m}}=0.048$). Validation gives $0.056$ against $0.057$ ($p=0.62$). The MathVista test results concern average accuracy and are unadjusted for multiple~\mbox{tests}.

\begin{table}[p]
\centering
\caption{\textbf{Both nulls across evaluation sets and thresholds.} Each set has $n$ prompts. The \emph{deg.} column counts prompts with all outcomes correct or all incorrect across the pool. These prompts contribute no permutation variance. The table reports observed gap $L$, each null mean, and upper- and lower-tail fractions. Upper tails count null statistics at least as large as observed, and lower tails count statistics at most as large. The split-half estimate selects both per-prompt members and the fixed comparator on $k/2$ responses and scores them on the other half, averaging $200$ splits. Each null uses $2{,}000$ samples. This is an independent run from Table~\ref{tab:null}, so permutation fractions differ slightly. The final block combines the two model~sizes.}
\label{tab:master}
\scriptsize
\renewcommand{\arraystretch}{1.0}
\setlength{\aboverulesep}{1pt}\setlength{\belowrulesep}{1.2pt}
\setlength{\tabcolsep}{3pt}
\begin{tabular}{@{}l S[table-format=3] S[table-format=2] S[table-format=1.2] S[table-format=1.3] S[table-format=1.3] S[table-format=1.3] S[table-format=1.3] S[table-format=1.3] S[table-format=1.3] S[table-format=1.3] S[table-format=+1.3]@{}}
\toprule
\textbf{Set} & {$n$} & {deg.} & {$\tau$} & {$L$} & {null} & {$p$} & {$p_{\downarrow}$} & {margin} & {$p_{\mathrm{m}}$} & {$p_{\mathrm{m}\downarrow}$} & {split-half} \\
\midrule
Geometry3K val & 240 & 39 & 0.05 & 0.088 & 0.103 & 0.959 & 0.083 & 0.101 & 0.954 & 0.104 & -0.021 \\
 & 240 & 39 & 0.10 & 0.067 & 0.105 & 1.000 & 0.001 & 0.103 & 0.999 & 0.005 & -0.019 \\
 & 240 & 39 & 0.20 & 0.108 & 0.100 & 0.303 & 0.814 & 0.097 & 0.230 & 0.855 & -0.000 \\
 & 240 & 39 & 0.30 & 0.096 & 0.090 & 0.374 & 0.755 & 0.088 & 0.312 & 0.802 & -0.008 \\
\midrule
Geometry3K test & 240 & 37 & 0.05 & 0.142 & 0.126 & 0.042 & 0.992 & 0.125 & 0.034 & 0.992 & +0.021 \\
 & 240 & 37 & 0.10 & 0.113 & 0.096 & 0.062 & 0.975 & 0.095 & 0.068 & 0.969 & +0.020 \\
 & 240 & 37 & 0.20 & 0.075 & 0.078 & 0.699 & 0.457 & 0.077 & 0.662 & 0.494 & +0.001 \\
 & 240 & 37 & 0.30 & 0.075 & 0.080 & 0.744 & 0.409 & 0.079 & 0.722 & 0.432 & -0.002 \\
\midrule
MathVista val & 137 & 42 & 0.50 & 0.029 & 0.018 & 0.151 & 0.970 & 0.017 & 0.127 & 0.980 & +0.009 \\
 & 137 & 42 & 0.60 & 0.044 & 0.044 & 0.679 & 0.639 & 0.042 & 0.579 & 0.729 & +0.009 \\
 & 137 & 42 & 0.70 & 0.080 & 0.081 & 0.645 & 0.572 & 0.078 & 0.541 & 0.670 & +0.010 \\
 & 137 & 42 & 0.80 & 0.102 & 0.099 & 0.523 & 0.687 & 0.096 & 0.448 & 0.741 & -0.012 \\
 & 137 & 42 & 0.90 & 0.036 & 0.065 & 0.992 & 0.030 & 0.065 & 0.986 & 0.046 & -0.045 \\
\midrule
MathVista test & 184 & 51 & 0.50 & 0.038 & 0.035 & 0.474 & 0.763 & 0.035 & 0.509 & 0.748 & +0.010 \\
 & 184 & 51 & 0.60 & 0.033 & 0.042 & 0.913 & 0.226 & 0.043 & 0.944 & 0.161 & +0.003 \\
 & 184 & 51 & 0.70 & 0.065 & 0.058 & 0.331 & 0.838 & 0.059 & 0.372 & 0.799 & +0.011 \\
 & 184 & 51 & 0.80 & 0.071 & 0.071 & 0.623 & 0.562 & 0.073 & 0.675 & 0.515 & -0.003 \\
 & 184 & 51 & 0.90 & 0.071 & 0.079 & 0.824 & 0.294 & 0.080 & 0.846 & 0.280 & -0.025 \\
\midrule
Geometry3K test, $k{=}64$ & 240 & 17 & 0.05 & 0.050 & 0.053 & 0.733 & 0.448 & 0.053 & 0.746 & 0.431 & +0.004 \\
 & 240 & 17 & 0.10 & 0.058 & 0.047 & 0.115 & 0.958 & 0.047 & 0.121 & 0.956 & +0.003 \\
 & 240 & 17 & 0.20 & 0.037 & 0.039 & 0.707 & 0.514 & 0.039 & 0.666 & 0.549 & -0.006 \\
 & 240 & 17 & 0.30 & 0.025 & 0.039 & 0.987 & 0.050 & 0.039 & 0.982 & 0.056 & -0.006 \\
\midrule
MathVista test, $k{=}64$ & 184 & 28 & 0.50 & 0.011 & 0.012 & 0.736 & 0.637 & 0.011 & 0.703 & 0.680 & +0.001 \\
 & 184 & 28 & 0.60 & 0.022 & 0.029 & 0.927 & 0.251 & 0.028 & 0.895 & 0.292 & -0.004 \\
 & 184 & 28 & 0.70 & 0.033 & 0.029 & 0.470 & 0.796 & 0.029 & 0.445 & 0.797 & -0.001 \\
 & 184 & 28 & 0.80 & 0.038 & 0.034 & 0.443 & 0.799 & 0.034 & 0.413 & 0.819 & +0.002 \\
 & 184 & 28 & 0.90 & 0.038 & 0.030 & 0.216 & 0.946 & 0.029 & 0.226 & 0.938 & -0.000 \\
\midrule
Geometry3K, $361$ new & 361 & 53 & 0.05 & 0.119 & 0.122 & 0.718 & 0.411 & 0.122 & 0.723 & 0.393 & -0.002 \\
 & 361 & 53 & 0.10 & 0.125 & 0.119 & 0.341 & 0.762 & 0.119 & 0.352 & 0.750 & -0.002 \\
 & 361 & 53 & 0.20 & 0.108 & 0.092 & 0.054 & 0.973 & 0.092 & 0.062 & 0.965 & +0.013 \\
 & 361 & 53 & 0.30 & 0.089 & 0.087 & 0.487 & 0.641 & 0.087 & 0.481 & 0.630 & +0.010 \\
\midrule
Geometry3K, $601$ pooled & 601 & 90 & 0.05 & 0.128 & 0.128 & 0.583 & 0.532 & 0.128 & 0.587 & 0.527 & +0.004 \\
 & 601 & 90 & 0.10 & 0.123 & 0.114 & 0.131 & 0.913 & 0.115 & 0.146 & 0.906 & +0.006 \\
 & 601 & 90 & 0.20 & 0.101 & 0.091 & 0.073 & 0.955 & 0.091 & 0.074 & 0.956 & +0.011 \\
 & 601 & 90 & 0.30 & 0.095 & 0.088 & 0.191 & 0.871 & 0.089 & 0.206 & 0.872 & +0.010 \\
\midrule
3B val & 240 & 45 & 0.05 & 0.133 & 0.127 & 0.336 & 0.818 & 0.126 & 0.338 & 0.808 & +0.006 \\
 & 240 & 45 & 0.10 & 0.138 & 0.134 & 0.475 & 0.657 & 0.134 & 0.467 & 0.674 & +0.006 \\
 & 240 & 45 & 0.20 & 0.104 & 0.116 & 0.868 & 0.203 & 0.116 & 0.876 & 0.212 & +0.008 \\
 & 240 & 45 & 0.30 & 0.100 & 0.101 & 0.602 & 0.537 & 0.101 & 0.622 & 0.517 & -0.001 \\
\midrule
3B test & 240 & 45 & 0.05 & 0.121 & 0.117 & 0.471 & 0.704 & 0.115 & 0.365 & 0.792 & +0.015 \\
 & 240 & 45 & 0.10 & 0.121 & 0.128 & 0.799 & 0.308 & 0.125 & 0.696 & 0.431 & +0.015 \\
 & 240 & 45 & 0.20 & 0.117 & 0.106 & 0.230 & 0.868 & 0.100 & 0.123 & 0.932 & -0.002 \\
 & 240 & 45 & 0.30 & 0.092 & 0.106 & 0.923 & 0.142 & 0.101 & 0.832 & 0.275 & -0.005 \\
\midrule
7B $+$ 3B test & 240 & 19 & 0.05 & 0.158 & 0.149 & 0.186 & 0.933 & 0.115 & {$<$0.001} & 1.000 & +0.037 \\
 & 240 & 19 & 0.10 & 0.150 & 0.148 & 0.531 & 0.624 & 0.103 & {$<$0.001} & 1.000 & +0.038 \\
 & 240 & 19 & 0.20 & 0.121 & 0.133 & 0.875 & 0.205 & 0.092 & 0.010 & 0.996 & +0.015 \\
 & 240 & 19 & 0.30 & 0.104 & 0.127 & 0.976 & 0.050 & 0.085 & 0.057 & 0.975 & +0.007 \\
\bottomrule
\end{tabular}
\end{table}

\paragraph{Overall coverage differences between members.} We measure the spread $\max_m\Cov_{P_m}(\tau)-\min_m\Cov_{P_m}(\tau)$ and compare it with a reference that independently permutes member labels within each prompt. This preserves each prompt's collection of estimated member rates while removing consistent member identity across prompts. It is different from re-dealing individual response outcomes, and stronger than the single common relabeling implied by an exchangeable training~\mbox{design}.

At the four ascending Geometry3K test thresholds, the spreads are $(0.021,\allowbreak 0.017,\allowbreak 0.046,\allowbreak 0.042)$. The reference $95$th percentiles are $(0.067,\allowbreak 0.058,\allowbreak 0.054,\allowbreak 0.054)$, with upper-tail fractions $(0.95,\allowbreak 0.96,\allowbreak 0.21,\allowbreak 0.28)$. Validation spreads are $(0.071,\allowbreak 0.079,\allowbreak 0.033,\allowbreak 0.042)$, with fractions $(0.021,\allowbreak 0.005,\allowbreak 0.60,\allowbreak 0.24)$. Seed $5$ leads at the low thresholds. No MathVista test threshold has $p<0.05$ ($p\ge0.24$), while validation at $\tau=0.90$ gives $p=0.004$. A realized leader is compatible with exchangeability of the training procedure and can affect the retrospectively selected best-member~\mbox{comparator}.

\section{Cross-dataset pool details}
\label{app:specialist}

The cross-dataset pools use the same base model for all members and vary the training dataset, under the light recipe (ten seeds, five per dataset) and the stronger recipe (four seeds, two per dataset; Appendix~\ref{app:protocol}). We evaluate every Geometry3K-trained seed on MathVista and every MathVista-trained seed on Geometry3K. Cross-evaluation follows the same procedure as the other comparisons, with $k=16$, a limit of $1{,}024$ new tokens, and the corrected verifier. Every member thus has $16$ responses on each of the $424$ union test prompts ($240$ Geometry3K and $184$ MathVista). The six thresholds span the two datasets' reliability bands. Coverage uses raw rate estimates, and both nulls follow~Appendix~\ref{app:audit}.

The dataset-identity router sends each prompt to the seed trained on its source dataset. The two embedding routers use the base model's image-and-text representations and train on the union of the validation sets ($377$ prompts), where both members also have $16$ responses per~\mbox{prompt}.

\paragraph{The two-member light pool.} Table~\ref{tab:specialist} reports the two-member light pool at all six thresholds. At $\tau=0.10$, disagreement is $D=0.120$, compared with plain-null mean $0.095$ ($p<0.001$, margin-preserving $p_{\mathrm{m}}=0.003$). The gap $L$, however, does not significantly exceed either null. The embedding linear router gains $-0.005$ with interval $[-0.017,+0.005]$ and disagreement error $0.41$. As reported in the main text, the Geometry3K-trained seed covers more prompts in both dataset halves at this~\mbox{threshold}.

\paragraph{The ten-member light pool.} All ten light seeds have complete cross-dataset evaluations. Their mean success rates range from $0.346$ to $0.377$ on Geometry3K and from $0.607$ to $0.633$ on MathVista. Thus changing the training dataset produces only modest separation in overall performance. Table~\ref{tab:specialist_all} tests whether the larger pool nevertheless has a reproducible oracle advantage. Only the lowest threshold shows a significant excess under either~\mbox{null}.

\paragraph{The stronger-recipe pools.} Table~\ref{tab:specialist_long} reports the two-member pool (seed $1$ of each dataset) and the four-member pool at every threshold. The routers are fitted on the union validation probes, where every stronger seed also has $16$ responses per prompt. At $\tau=0.10$ the Geometry3K-trained member covers $0.650$ of Geometry3K prompts and $0.832$ of MathVista prompts, against $0.629$ and $0.810$ for the MathVista-trained member. At $\tau=0.70$ the figures are $0.375$ and $0.582$ against $0.271$ and $0.625$, so the members separate on MathVista prompts only at high thresholds, which is where the routers gain. Across the $36$ router intervals of the two pools, one lies wholly above zero and none wholly below. Against the member selected on validation the two-member gains are unchanged, since validation and test select the same member at every threshold but $0.05$. On the four-member pool validation selects a different seed at several thresholds, including the other Geometry3K seed at $\tau=0.70$, where the gains rise to $+0.038$ for the identity router and $+0.057$ for the nearest-neighbor~\mbox{router}.

\begin{center}
\begin{minipage}{\textwidth}
\centering
\captionof{table}{\textbf{The two-member light cross-dataset pool on $424$ test prompts.} The first block compares the observed oracle gap $L$ with the plain and margin-preserving nulls. The next block reports estimated coverage. Router gains use the best member at each threshold as the comparator. \emph{Identity} sends a prompt to the policy trained on its dataset. \emph{Nearest} uses the embedding nearest-neighbor router. The final column is the latter's $95\%$ prompt-bootstrap interval. Mixture coverage uses all pooled responses. Table~\ref{tab:main} gives the budget-matched~estimate.}
\label{tab:specialist}
\scriptsize
\setlength{\tabcolsep}{2.4pt}
\begin{tabular}{@{}S[table-format=1.2] S[table-format=1.3] S[table-format=1.3] S[table-format=1.3] S[table-format=1.3] S[table-format=1.3] S[table-format=1.3] S[table-format=1.3] S[table-format=1.3] S[table-format=+1.3] S[table-format=+1.3] c@{}}
\toprule
& \multicolumn{5}{c}{\textbf{$L(\tau)$}} & \multicolumn{3}{c}{\textbf{coverage}} & \multicolumn{3}{c}{\textbf{router gain}} \\
\cmidrule(lr){2-6}\cmidrule(lr){7-9}\cmidrule(lr){10-12}
{$\tau$} & {obs.} & {null} & {$p$} & {margin} & {$p_{\mathrm{m}}$} & {best} & {mixture} & {oracle} & {identity} & {nearest} & {$95\%$ interval} \\
\midrule
0.05 & 0.054 & 0.045 & 0.050 & 0.043 & 0.045 & 0.814 & 0.797 & 0.868 & -0.002 & +0.007 & $[-0.019,+0.033]$ \\
0.10 & 0.045 & 0.041 & 0.352 & 0.038 & 0.220 & 0.741 & 0.719 & 0.785 & -0.005 & -0.005 & $[-0.024,+0.014]$ \\
0.20 & 0.017 & 0.031 & 0.993 & 0.027 & 0.968 & 0.653 & 0.639 & 0.670 & -0.007 & -0.028 & $[-0.050,-0.009]$ \\
0.30 & 0.028 & 0.033 & 0.843 & 0.027 & 0.522 & 0.608 & 0.585 & 0.637 & -0.002 & -0.019 & $[-0.040,+0.002]$ \\
0.50 & 0.033 & 0.032 & 0.547 & 0.026 & 0.165 & 0.488 & 0.481 & 0.521 & -0.005 & -0.012 & $[-0.028,+0.002]$ \\
0.70 & 0.024 & 0.030 & 0.920 & 0.025 & 0.677 & 0.361 & 0.354 & 0.384 & +0.005 & -0.005 & $[-0.021,+0.014]$ \\
\bottomrule
\end{tabular}
\end{minipage}
\end{center}

\begin{center}
\begin{minipage}{\textwidth}
\centering
\captionof{table}{\textbf{The ten-member light cross-dataset pool on the union test set.} The table reports observed oracle gap $L$ and disagreement $D$, followed by null means and upper-tail fractions for $L$. The final columns give coverage for the best member, uniform mixture, and plug-in oracle. Mixture coverage averages all raw member rate estimates and therefore uses more responses than a single~member.}
\label{tab:specialist_all}
\scriptsize
\setlength{\tabcolsep}{3.5pt}
\begin{tabular}{@{}S[table-format=1.2] S[table-format=1.3] S[table-format=1.3] S[table-format=1.3] S[table-format=1.3] S[table-format=1.3] S[table-format=1.3] S[table-format=1.3] S[table-format=1.3] S[table-format=1.3]@{}}
\toprule
{$\tau$} & {$L$} & {$D$} & {null} & {$p$} & {margin} & {$p_{\mathrm{m}}$} & {best} & {mixture} & {oracle} \\
\midrule
0.05 & 0.123 & 0.288 & 0.108 & {$<$0.001} & 0.107 & 0.001 & 0.814 & 0.807 & 0.936 \\
0.10 & 0.104 & 0.238 & 0.100 & 0.344 & 0.098 & 0.287 & 0.741 & 0.750 & 0.844 \\
0.20 & 0.087 & 0.231 & 0.103 & 0.979 & 0.101 & 0.965 & 0.653 & 0.642 & 0.741 \\
0.30 & 0.094 & 0.222 & 0.096 & 0.662 & 0.094 & 0.555 & 0.608 & 0.568 & 0.703 \\
0.50 & 0.092 & 0.208 & 0.082 & 0.096 & 0.080 & 0.064 & 0.493 & 0.481 & 0.585 \\
0.70 & 0.078 & 0.186 & 0.084 & 0.829 & 0.082 & 0.756 & 0.370 & 0.373 & 0.448 \\
\bottomrule
\end{tabular}
\end{minipage}
\end{center}

\begin{center}
\begin{minipage}{\textwidth}
\centering
\captionof{table}{\textbf{The stronger-recipe cross-dataset pools on the union test set.} Observed oracle gap $L$ and disagreement $D$, then the plain and margin-preserving null means for $L$ with upper-tail fractions, the coverage of the best member, the uniform mixture and the plug-in oracle, and the coverage gain over the best member of the dataset-identity router and of the embedding nearest-neighbor router, with the latter's $95\%$ prompt-bootstrap interval. Mixture coverage uses all pooled~responses.}
\label{tab:specialist_long}
\scriptsize
\setlength{\tabcolsep}{2.2pt}
\begin{tabular}{@{}S[table-format=1.2] S[table-format=1.3] S[table-format=1.3] S[table-format=1.3] S[table-format=1.3] S[table-format=1.3] S[table-format=1.3] S[table-format=1.3] S[table-format=1.3] S[table-format=1.3] S[table-format=+1.3] S[table-format=+1.3] c@{}}
\toprule
& \multicolumn{2}{c}{\textbf{gap}} & \multicolumn{4}{c}{\textbf{nulls for $L$}} & \multicolumn{3}{c}{\textbf{coverage}} & \multicolumn{3}{c}{\textbf{router gain}} \\
\cmidrule(lr){2-3}\cmidrule(lr){4-7}\cmidrule(lr){8-10}\cmidrule(lr){11-13}
{$\tau$} & {$L$} & {$D$} & {null} & {$p$} & {margin} & {$p_{\mathrm{m}}$} & {best} & {mixture} & {oracle} & {identity} & {nearest} & {$95\%$ interval} \\
\midrule
\multicolumn{13}{@{}l}{\emph{two members (seed $1$ of each dataset)}} \\
\midrule
0.05 & 0.059 & 0.120 & 0.039 & {$<$0.001} & 0.036 & {$<$0.001} & 0.797 & 0.802 & 0.856 & -0.007 & -0.009 & $[-0.035,+0.017]$ \\
0.10 & 0.054 & 0.130 & 0.032 & {$<$0.001} & 0.028 & {$<$0.001} & 0.729 & 0.724 & 0.783 & -0.009 & -0.009 & $[-0.028,+0.009]$ \\
0.20 & 0.035 & 0.111 & 0.030 & 0.230 & 0.025 & 0.063 & 0.660 & 0.663 & 0.696 & -0.014 & -0.012 & $[-0.028,+0.005]$ \\
0.30 & 0.047 & 0.127 & 0.027 & {$<$0.001} & 0.022 & {$<$0.001} & 0.627 & 0.606 & 0.675 & -0.014 & -0.014 & $[-0.031,+0.000]$ \\
0.50 & 0.047 & 0.116 & 0.022 & {$<$0.001} & 0.017 & {$<$0.001} & 0.535 & 0.526 & 0.583 & +0.009 & +0.005 & $[-0.014,+0.024]$ \\
0.70 & 0.040 & 0.120 & 0.026 & 0.002 & 0.021 & {$<$0.001} & 0.465 & 0.443 & 0.505 & +0.019 & +0.014 & $[-0.005,+0.033]$ \\
\midrule
\multicolumn{13}{@{}l}{\emph{four members}} \\
\midrule
0.05 & 0.097 & 0.222 & 0.068 & {$<$0.001} & 0.064 & {$<$0.001} & 0.797 & 0.788 & 0.894 & -0.007 & -0.026 & $[-0.061,+0.009]$ \\
0.10 & 0.092 & 0.205 & 0.062 & {$<$0.001} & 0.057 & {$<$0.001} & 0.731 & 0.731 & 0.823 & -0.012 & -0.007 & $[-0.033,+0.019]$ \\
0.20 & 0.073 & 0.196 & 0.050 & {$<$0.001} & 0.044 & {$<$0.001} & 0.660 & 0.672 & 0.733 & -0.014 & -0.005 & $[-0.031,+0.024]$ \\
0.30 & 0.083 & 0.215 & 0.049 & {$<$0.001} & 0.042 & {$<$0.001} & 0.627 & 0.611 & 0.710 & -0.014 & +0.012 & $[-0.014,+0.040]$ \\
0.50 & 0.075 & 0.205 & 0.046 & {$<$0.001} & 0.038 & {$<$0.001} & 0.550 & 0.531 & 0.625 & -0.005 & +0.021 & $[-0.002,+0.045]$ \\
0.70 & 0.085 & 0.217 & 0.048 & {$<$0.001} & 0.042 & {$<$0.001} & 0.465 & 0.439 & 0.550 & +0.019 & +0.038 & $[+0.005,+0.068]$ \\
\bottomrule
\end{tabular}
\end{minipage}
\end{center}

\section{Voting within and across policies}
\label{app:vote}

Self-consistency returns the most frequent extracted answer among several responses from one policy \citep{wang2023selfconsistency}. We compare it with a pooled vote that draws a policy uniformly for each response, using vote counts $b\in\{1,3,5,7,9\}$ to check whether any benefit depends on the voting budget. Success probabilities are estimated by repeatedly sampling from stored responses. Ties are broken uniformly at random, and responses without an extractable answer abstain. These estimates and their single-response baselines both use raw~\mbox{frequencies}.

\begin{table}[!ht]
\centering
\caption{\textbf{Voting within and across policies.} Entries are rAUC on $[0.05,0.30]$ for Geometry3K and $[0.50,0.90]$ for MathVista. A pooled vote samples a policy uniformly for each of its $b$ responses. The last two columns compare pooled voting with the best member using one response and that same member using $b$ responses,~respectively.}
\label{tab:vote}
\small
\setlength{\tabcolsep}{4.5pt}
\begin{tabular}{@{}lllrrrr@{}}
\toprule
\textbf{Dataset} & \textbf{Split} & \textbf{$b$} & \textbf{Pooled vote} & \textbf{Best member's vote} & \textbf{Gain over single} & \textbf{Pooled $-$ member} \\
\midrule
Geometry3K & val & 1 & $0.553$ & $0.566$ & $-0.013$ & $-0.013$ \\
Geometry3K & val & 3 & $0.549$ & $0.556$ & $-0.017$ & $-0.007$ \\
Geometry3K & val & 5 & $0.557$ & $0.553$ & $-0.009$ & $+0.004$ \\
Geometry3K & val & 7 & $0.561$ & $0.549$ & $-0.005$ & $+0.012$ \\
Geometry3K & val & 9 & $0.560$ & $0.546$ & $-0.006$ & $+0.014$ \\
Geometry3K & test & 1 & $0.584$ & $0.595$ & $-0.011$ & $-0.011$ \\
Geometry3K & test & 3 & $0.587$ & $0.594$ & $-0.008$ & $-0.007$ \\
Geometry3K & test & 5 & $0.602$ & $0.599$ & $+0.008$ & $+0.004$ \\
Geometry3K & test & 7 & $0.609$ & $0.600$ & $+0.015$ & $+0.009$ \\
Geometry3K & test & 9 & $0.612$ & $0.600$ & $+0.017$ & $+0.011$ \\
\midrule
MathVista & val & 1 & $0.605$ & $0.619$ & $-0.014$ & $-0.014$ \\
MathVista & val & 3 & $0.668$ & $0.666$ & $+0.049$ & $+0.002$ \\
MathVista & val & 5 & $0.698$ & $0.691$ & $+0.079$ & $+0.007$ \\
MathVista & val & 7 & $0.710$ & $0.702$ & $+0.091$ & $+0.008$ \\
MathVista & val & 9 & $0.717$ & $0.708$ & $+0.098$ & $+0.009$ \\
MathVista & test & 1 & $0.539$ & $0.546$ & $-0.008$ & $-0.008$ \\
MathVista & test & 3 & $0.587$ & $0.588$ & $+0.041$ & $-0.001$ \\
MathVista & test & 5 & $0.604$ & $0.608$ & $+0.058$ & $-0.004$ \\
MathVista & test & 7 & $0.612$ & $0.622$ & $+0.065$ & $-0.010$ \\
MathVista & test & 9 & $0.616$ & $0.632$ & $+0.070$ & $-0.015$ \\
\bottomrule
\end{tabular}
\end{table}
On Geometry3K validation, pooled voting remains below the best single-response policy at every vote count. On test, it exceeds that member's own vote from $b=5$ onward, but the differences are not statistically resolved. At $b=9$, the $95\%$ interval is $[-0.013,+0.035]$ for the gain over the budget-matched single-member vote and $[-0.007,+0.042]$ for the gain over one response. These results do not establish a benefit from voting across~\mbox{policies}.

\section{Replications}
\label{app:replications}

\subsection{MathVista}
\label{app:ds2}

The MathVista study follows the primary study's metrics and verifier. Dataset eligibility, split assignment, and the reliability-band selection rule were fixed before scoring test~\mbox{responses}.

\paragraph{Dataset and band selection.} We considered MathVerse \citep{zhang2024mathverse} and MathVista for the second study. Before scoring either candidate's test split, we fixed a two-stage rule and applied it to base-model validation samples of $16$ responses per prompt. We exclude overlap with Geometry3K, including the GEOS source subset of MathVista. MathVerse prompts are grouped by source problem so that its five variants cannot cross~\mbox{splits}.

The first stage requires at least half the validation prompts to yield both successes and failures, with between one and fifteen correct responses out of sixteen. MathVerse fails this criterion and MathVista passes (Table~\ref{tab:ds2cal}). The second stage retains $[0.05,0.30]$ unless one of seven prespecified bands contains at least five percentage points more of these mixed-outcome prompts. For MathVista, $[0.50,0.90]$ contains the largest share, $0.465$, compared with $0.174$ in the original band. We therefore use $[0.50,0.90]$ and report thresholds~$\{0.5,0.6,0.7,0.8,0.9\}$.

\begin{table}[!htbp]
\centering
\caption{\textbf{Dataset selection using base-model validation responses.} Each prompt has $16$ sampled responses. The two extreme-count columns give fractions of all prompts. \emph{Mixed outcomes} is the fraction with both correct and incorrect responses. \emph{Band share} is the fraction within that mixed-outcome group whose estimated success rate lies in~$[0.05,0.30]$.}
\label{tab:ds2cal}
\small
\setlength{\tabcolsep}{3pt}
\begin{tabular}{@{}lrrrrrr@{}}
\toprule
\textbf{Candidate} & \textbf{Items} & \textbf{Mean rate} & \textbf{0/16 correct} & \textbf{16/16 correct} & \textbf{Mixed outcomes} & \textbf{Band share} \\
\midrule
MathVerse & $240$ & $0.271$ & $0.446$ & $0.067$ & $0.487$ & $0.496$ \\
MathVista & $137$ & $0.638$ & $0.117$ & $0.255$ & $\mathbf{0.628}$ & $0.174$ \\
\bottomrule
\end{tabular}
\end{table}
\paragraph{Splits and evaluation.} After exclusions, MathVista contributes $916$ public items, assigned by hash to $366$ training, $229$ reserved, $137$ validation, and $184$ test items. Five seeds follow the primary training recipe and are evaluated with $16$ responses per prompt. No test threshold rejects the member-spread diagnostic at level $0.05$. Because $M\tau>1$ throughout the selected band, the uniform-mixture lower bound is zero and gives no useful guarantee here. Table~\ref{tab:main} reports the deployment~\mbox{comparisons}.

\subsection{Budget, prompts, and backbone}

We repeat the primary analysis with more responses, more prompts, and a smaller base model. Table~\ref{tab:replications} reports system performance. Table~\ref{tab:master} gives both nulls and split-half estimates at each~\mbox{threshold}.

\paragraph{The three extensions.} First, we increase the evaluation budget from $16$ to $64$ responses per member and prompt on both test sets. Second, we evaluate the $361$ remaining Geometry3K test prompts, both separately and together with the original $240$. Third, we repeat training with five seeds of Qwen2.5-VL-3B-Instruct, changing only the model size. Before test scoring, base-model validation rates select the band $[0.05,0.30]$, which contains $0.618$ of validation prompts with mixed outcomes. Routers for the new Geometry3K prompts reuse the primary validation fit. The 3B routers are fitted on 3B validation~\mbox{outcomes}.

\begin{table}[!ht]
\centering
\caption{\textbf{Replication results for mixtures and routers.} Entries are rAUC on each dataset's selected band, with $k=16$ unless stated otherwise. Pools contain the trained policies. The base is reported separately. The mixture uses the same response budget as a member. The best member maximizes rAUC on the evaluated split. Dashes omit routing results on 3B validation, which is used for router fitting. The final two rows summarize threshold-specific routing intervals and the member-spread~diagnostic.}
\label{tab:replications}
\scriptsize
\setlength{\tabcolsep}{1pt}
\resizebox{\linewidth}{!}{\begin{tabular}{@{}l *{6}{S[table-format=+1.4]}@{}}
\toprule
\textbf{System or quantity} & {\bfseries Geometry3K} & {\bfseries MathVista} & {\bfseries Geometry3K} & {\bfseries Geometry3K} & {\bfseries 3B backbone} & {\bfseries 3B backbone} \\
& {\bfseries test, $k{=}64$} & {\bfseries test, $k{=}64$} & {\bfseries $361$ new, test} & {\bfseries $601$ pooled, test} & {\bfseries validation} & {\bfseries test} \\
& {\bfseries $240$ prompts} & {\bfseries $184$ prompts} & {\bfseries $361$ prompts} & {\bfseries $601$ prompts} & {\bfseries $240$ prompts} & {\bfseries $240$ prompts} \\
\midrule
base $B$ & 0.562 & 0.509 & 0.531 & 0.550 & 0.382 & 0.371 \\
best member by rAUC & 0.598 & 0.549 & 0.550 & 0.567 & 0.446 & 0.482 \\
oracle, best member per prompt & 0.644 & 0.581 & 0.666 & 0.681 & 0.576 & 0.601 \\
mixture, uniform over the seeds & 0.584 & 0.539 & 0.546 & 0.564 & 0.436 & 0.459 \\
routing, best family, no test labels & 0.590 & 0.547 & 0.543 & 0.566 & {--} & 0.480 \\
\midrule
routing cells with interval below zero & {0 of 12} & {3 of 15} & {7 of 12} & {5 of 12} & {--} & {2 of 12} \\
member-spread test, smallest $p$ over thresholds & 0.076 & 0.065 & 0.42 & 0.39 & 0.35 & 0.002 \\
\bottomrule
\end{tabular}}
\end{table}

\paragraph{Oracle gaps.} None of the $25$ replication conditions significantly exceeds the plain re-deal null for $L$ at level $0.05$ (smallest $p=0.054$, Table~\ref{tab:master}). Of the $25$ gaps, $24$ lie within or on the boundary of the null's $90\%$ range and one lies below it. Matched split-half estimates range from $-0.006$ to $+0.015$. Growth in the gap with pool size exceeds its null once, on 3B test at $\tau=0.05$ ($0.064$ versus $0.054$, unadjusted~$p=0.033$).

\paragraph{Mixtures.} At a matched response budget, the mixture trails the best member on every replication set. At $k=64$, its rAUC deficits are $0.014$ on Geometry3K, with interval $[-0.023,-0.004]$, and $0.010$ on MathVista, with interval $[-0.018,-0.002]$. The deficits on the $361$ new and $601$ pooled Geometry3K prompts are $0.004$ and $0.003$, with intervals $[-0.020,+0.014]$ and $[-0.015,+0.009]$. On 3B test, the deficit is $0.024$ from unrounded values, with interval $[-0.045,-0.003]$. Against the validation-selected 3B member, however, the difference is $+0.001$ with interval $[-0.023,+0.023]$. The resolved $k=64$ disadvantage applies to comparison with the retrospectively selected best~\mbox{member}.

\paragraph{Routing and member differences.} None of the $63$ routing intervals lies wholly above zero, while $17$ lie below it. Estimated disagreement error falls below the break-even value in only two conditions, each with gain $+0.005$, and equals it in four others, with gain $0.000$. The member-spread diagnostic has no $p<0.05$ at $k=64$, on either Geometry3K prompt extension, or on 3B validation. On 3B test, seed $3$ leads at $\tau=0.10$ and $0.20$, with unadjusted $p=0.002$ and~$0.023$.

\paragraph{Disagreement without a resolved gain.} At $\tau=0.20$, disagreement exceeds its plain-null mean on the $361$ new Geometry3K prompts ($0.233$ versus $0.209$, $p=0.023$), the pooled $601$ prompts ($0.223$ versus $0.204$, $p=0.018$), and 3B test ($0.287$ versus $0.254$, $p=0.015$). The oracle gaps are near the upper end of the null range on the two Geometry3K sets and within it on 3B test. Their matched split-half gaps are $+0.013$, $+0.011$, and $-0.002$. The tested routers recover no statistically resolved benefit on these~\mbox{sets}.

\section{Verifier correction}
\label{app:integrity}

The same correctness checker supplies training rewards and scores evaluation responses, so a parsing defect can influence both what policies learn and how they are measured. Our audit found that answer normalization failed to remove a typeset degree mark. An answer such as $62^\circ$ was marked incorrect against a reference answer of~$62$.

Re-scoring the audited cache of $57{,}600$ responses ($15$ arms of $240$ Geometry3K prompts with $16$ responses each: the base model and the five light seeds on both splits, two training checkpoints on validation, and one superseded shorter-budget base run that the analysis excludes) with the corrected checker changed $237$ verdicts, all from incorrect to correct. All results in this manuscript use the corrected evaluation checker. A policy can change answer formatting during training. The same parsing defect can therefore affect trained and base policies differently. The Geometry3K light seeds were trained before the correction, so the defective checker supplied their rewards. The MathVista seeds, the 3B seeds, and the stronger recipe were trained with the corrected checker, and every evaluation uses~it.

\end{document}

%% file: affiliations.tex
\DeclareAffiliation{hbku}{%
  College of Science and Engineering, Hamad Bin Khalifa University, Doha, Qatar}

\DeclareAffiliation{tamu}{%
  Department of Computer and Electrical Engineering,
  Texas A\&M University, College Station, TX, USA}

\DeclareAffiliation{iub}{%
  Luddy School of Informatics, Computing, and Engineering,
  Indiana University Bloomington, Bloomington, IN, USA}
